\documentclass[lettersize,journal]{IEEEtran}
\usepackage{amsmath,amsfonts}
\usepackage{array}
\usepackage{color, colortbl}
\usepackage[caption=false,font=normalsize,labelfont=sf,textfont=sf]{subfig}
\usepackage{textcomp}
\usepackage{stfloats}
\usepackage{url}
\usepackage{verbatim}
\usepackage{graphicx}
\usepackage{cite}
\definecolor{babyblue}{rgb}{0.54, 0.81, 0.94}
\newcommand*{\belowrulesepcolor}[1]{%
  \noalign{%
    \kern-\belowrulesep 
    \begingroup 
      \color{#1}%
      \hrule height\belowrulesep 
    \endgroup 
  }%
}

\newcommand*{\aboverulesepcolor}[1]{%
  \noalign{%
    \begingroup 
      \color{#1}%
      \hrule height\aboverulesep 
    \endgroup 
    \kern-\aboverulesep 
  }%
}

\usepackage{multirow}
\usepackage{bbding}
\usepackage{booktabs}
\usepackage{hyperref}
\usepackage{float}
\usepackage[ruled,linesnumbered]{algorithm2e}
\usepackage{xcolor}
\usepackage{amsthm,amsmath,amssymb}
\usepackage{mathrsfs}
\usepackage{orcidlink}
\usepackage{lineno}

\begin{document}

\title{TOPIC: A Parallel Association Paradigm for Multi-Object Tracking under Complex Motions and Diverse Scenes}
\author{Xiaoyan Cao\orcidlink{0000-0002-4980-2050}*,
        Yiyao Zheng\orcidlink{0000-0001-5041-8630}*,
        Yao Yao\orcidlink{0000-0001-9887-4301}*,
        Huapeng Qin\orcidlink{0000-0002-5132-5329},
        Xiaoyu Cao\orcidlink{0000-0003-0219-1798},
        Shihui Guo\orcidlink{0000-0002-1473-297X},~\IEEEmembership{Senior Member, IEEE}
        
\thanks{
\textit{*Equal contribution. Corresponding author: Shihui Guo.}}
\thanks{
Xiaoyan Cao and Huapeng Qin are with Key Laboratory for Urban Habitat Environmental Science and Technology, School of Environment and Energy, Peking University Shenzhen Graduate School, Shenzhen, Guangdong, China (e-mail: caoxiaoyan@stu.pku.edu.cn; qinhp@pkusz.edu.cn).}
\thanks{
Yiyao Zheng is with Quanzhou University of Information Engineering, Quanzhou, Fujian, China (e-mail: zyy112120@gmail.com).}
\thanks{
Yao Yao is with Tsinghua-Berkeley Shenzhen Institute, Tsinghua University, Shenzhen, Guangdong, China (e-mail: yaoyao19950630@gmail.com).
}
\thanks{Xiaoyu Cao and Shihui Guo are with the College of Chemistry and Chemical Engineering and the School of Informatics, Xiamen University, Xiamen, Fujian, China (e-mail: xcao@xmu.edu.cn; guoshihui@xmu.edu.cn).}
\thanks{This paper has supplementary downloadable material available at http://ieeexplore.ieee.org., provided by the author. The material includes Supplementary Material. Contact guoshihui@xmu.edu.cn for further questions about this work.
}
}%
\markboth{Journal of \LaTeX\ Class Files,~Vol.~14, No.~8, August~2021}%
{Shell \MakeLowercase{\textit{et al.}}: A Sample Article Using IEEEtran.cls for IEEE Journals}

\maketitle

\begin{figure*}[ht!]
\setlength{\abovecaptionskip}{0.2pt}
\setlength{\belowcaptionskip}{0.2pt}
\centering
\includegraphics[width=0.85\linewidth]{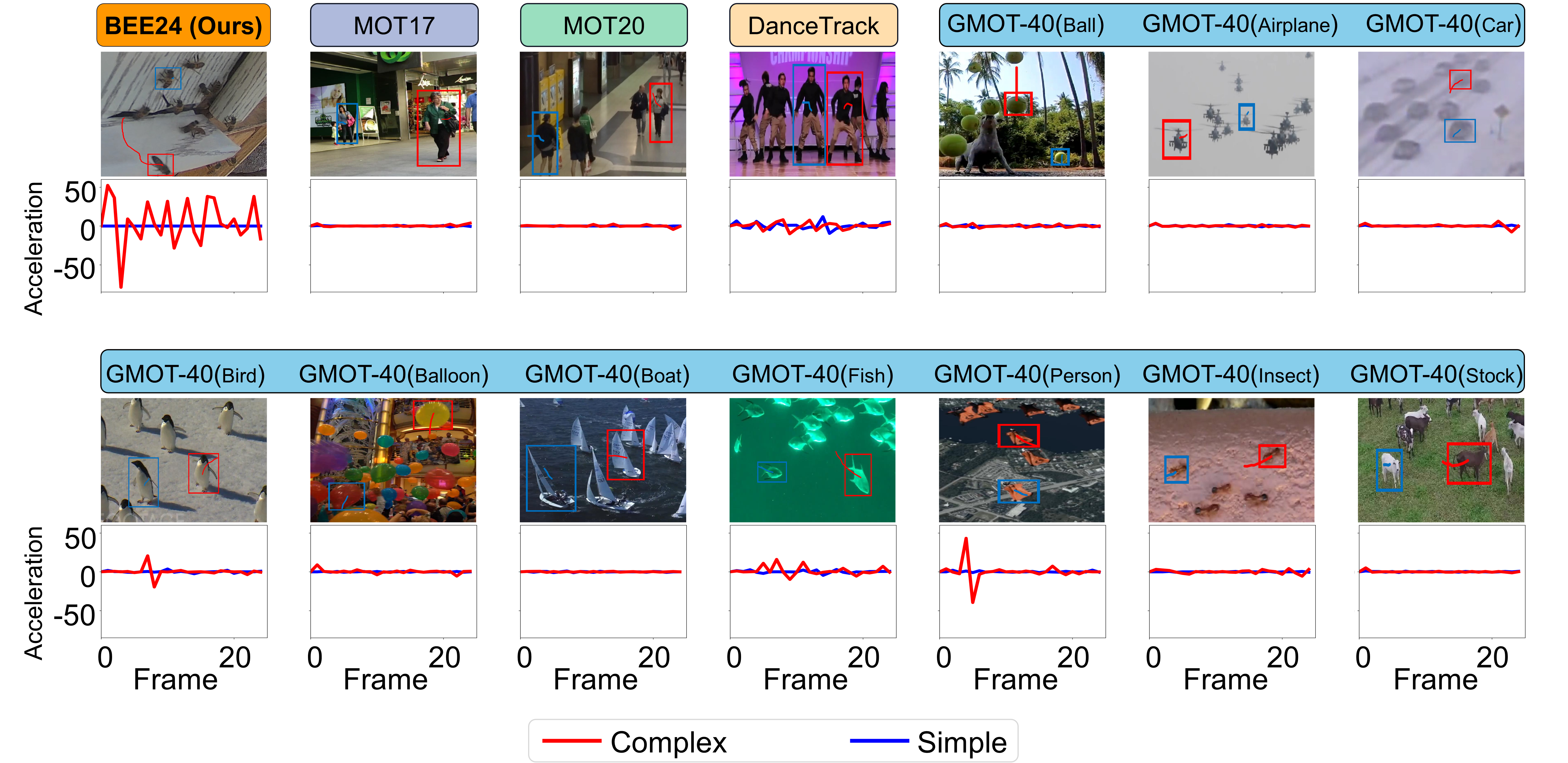}
\vspace{-0.1cm}
\caption{
    Comparison of the properties of different datasets.
    In addition to the properties of occlusion and highly similar appearance, the property of complex motion patterns is remarkable in BEE24.
    This can be seen in the diversity of motion patterns between objects and the variability of motion patterns of a single object.
    In the legend, ``Complex'' and ``Simple'' denote the objects with the most complex and simplest motion patterns in the scene, respectively.
   }
   \label{fig:data_motion}
\end{figure*}

\begin{figure*}[!h]
\setlength{\abovecaptionskip}{1pt}
\setlength{\belowcaptionskip}{1pt}
    \centering
    \includegraphics[width=0.85\linewidth]{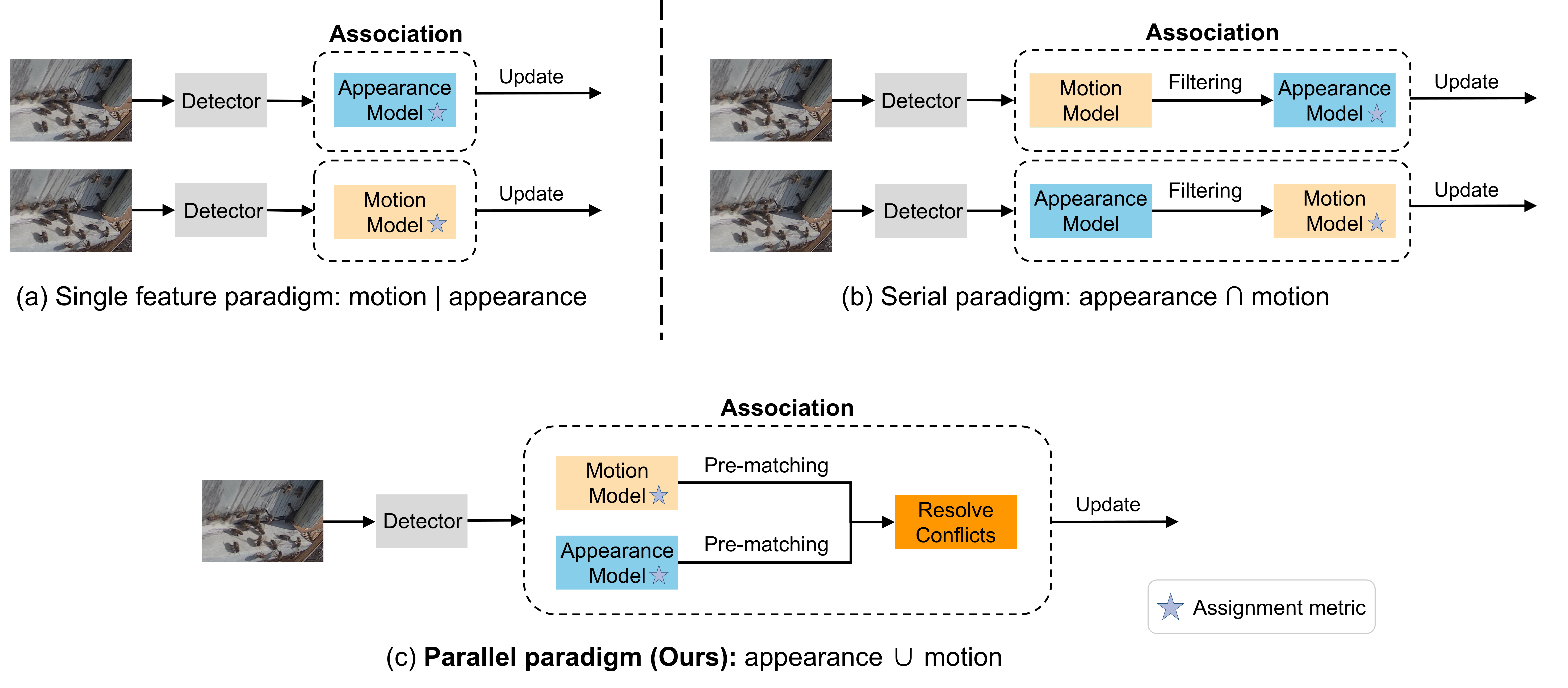}
    \caption{Comparison of existing association paradigms with our proposed parallel paradigm.
    (a) the single-feature association paradigm, either uses motion or appearance feature as assignment metric;
    (b) the serial association paradigm, manually specifies a feature to filter association candidates, followed by another feature as the primary assignment metric, which resembles taking the ``intersection'' of motion and appearance matches;
    (c) our proposed parallel association paradigm, uses motion and appearance features as assignment metrics in parallel, like taking the union set, and can resolve conflicts.}
    \label{fig:pipeline}
\end{figure*}

\begin{abstract}
Video data and algorithms have been driving advances in multi-object tracking (MOT). While existing MOT datasets focus on occlusion and appearance similarity, complex motion patterns are widespread yet overlooked. To address this issue, we introduce a new dataset called BEE24 to highlight complex motions. 
Identity association algorithms have long been the focus of MOT research.
Existing trackers can be categorized into two association paradigms: single-feature paradigm (based on either motion or appearance feature) and serial paradigm (one feature serves as secondary while the other is primary).
However, these paradigms are incapable of fully utilizing different features.
In this paper, we propose a parallel paradigm and present the Two rOund Parallel matchIng meChanism (TOPIC) to implement it.
The TOPIC leverages both motion and appearance features and can adaptively select the preferable one as the assignment metric based on motion level. 
Moreover, we provide an Attention-based Appearance Reconstruction Module (AARM) to reconstruct appearance feature embeddings, thus enhancing the representation of appearance features.
Comprehensive experiments show that our approach achieves state-of-the-art performance on four public datasets and BEE24.
Moreover, BEE24 challenges existing trackers to track multiple similar-appearing small objects with complex motions over long periods, which is critical in real-world applications such as beekeeping and drone swarm surveillance.
Notably, our proposed parallel paradigm surpasses the performance of existing association paradigms by a large margin, e.g., reducing false negatives by 6\% to 81\% compared to the single-feature association paradigm.
The introduced dataset and association paradigm in this work offer a fresh perspective for advancing the MOT field.
The source code and dataset are available at \url{https://github.com/holmescao/TOPICTrack}.
\end{abstract}


\section{Introduction}

\IEEEPARstart{M}{ulti}-object tracking (MOT) is a vital subfield in computer vision, which covers numerous applications,
such as robot navigation~\cite{xiang2015learning}, intelligent monitor~\cite{gomez2021smartmot}, and human-computer interaction~\cite{candamo2009understanding}.
In MOT, the task involves detecting the locations of objects of interest and associating their identities across frames in a given video~\cite{luo2021multiple}.
Dataset construction and algorithm optimization are two crucial aspects that empower trackers to cope with complex and diverse scenes.

From the dataset perspective, researchers commonly adopt two main approaches: (1) expanding diversity of scenes or object classes, as demonstrated by GMOT-40~\cite{bai2021gmot}; 
(2) enriching data properties by focusing on challenges like occlusion (e.g., MOT17~\cite{milan2016mot16}, MOT20~\cite{dendorfer2020mot20}) or high appearance similarity (e.g., DanceTrack~\cite{sun2022dancetrack}).
However, as depicted in Figure~\ref{fig:data_motion}, one key limitation in existing datasets is the relatively simplistic motion patterns. 
Specifically, different objects move similarly, with individual motions having low intensity and little variability across frames (e.g., MOT17, MOT20). 
In contrast, complex variable motion is common in life and nature, such as bee colonies around a hive~\cite{srinivasan1999motion}. 
Accurately tracking such complex motions is crucial in practical applications like beekeeping monitoring, analysis of bee colony behavior, and drone swarm surveillance.
To explore MOT in complex scenarios and address these real-world challenges, we build the BEE24 dataset exhibiting complex motion patterns in two key aspects: (1) diversity of motion among different objects within frames; (2) significant variability of individual object motions across frames (as shown in the 1st example in Figure~\ref{fig:data_motion}). We believe this dataset enriched with complex and diverse data properties, can serve as a more challenging benchmark for advancing general MOT research and improving the performance of tracking systems in practical applications.

From an algorithm optimization perspective, most trackers~\cite{li2019multiple, wang2020towards, zhang2021fairmot, wu2021track} since DeepSORT~\cite{wojke2017simple} in 2016 follows its serial association paradigm. 
This paradigm utilizes a feature to initially filter some association candidates, such as the appearance feature in TraDeS~\cite{li2019multiple} and the motion feature in FairMOT~\cite{zhang2021fairmot}, followed by using the other feature as the primary association metric, aiming to avoid the matching conflict problem that may arise from two features.
Such a paradigm resembles ``intersection'', which does not take full advantage of both features and may even be damaging to tracking performance, due to filtering may cause missing tracking, i.e., false negatives (FN).
Recent works like ByteTrack~\cite{zhang2022bytetrack} and OC-SORT~\cite{cao2023observation} using just motion feature, a typical single-feature association paradigm (as shown in Figure~\ref{fig:pipeline}(a)), outperform the two-feature serial association paradigms like FairMOT~\cite{zhang2021fairmot}. 
However, comparisons may be unfair due to detector differences. For example, OC-SORT uses the state-of-the-art YOLOX~\cite{ge2021yolox} detector while FairMOT uses the weaker DLA~\cite{yu2018deep}, impacting associations. 

Considering paradigms alone, intuitively employing more features should improve performance. Our focus is effectively combining features to maximize strengths.
To begin, we analyze feature performance to reveal insights. 
In low-speed scenes like MOT17 and DanceTrack, occlusion and high appearance similarity challenge the appearance feature. 
Here, motion features are more effective as the assignment metric due to simpler motions (Figure~\ref{fig:data_motion}). 
In high-speed scenes like BEE24 and GMOT-40 (Person), complex nonlinear motions pose a great challenge to motion models based on linear motion assumption. 
However, organisms commonly avoid collisions at high speeds by maintaining distance~\cite{richardson2017swarm}. This makes the appearance more visible, improving the distinguishability. Thus the appearance feature excels here.

Based on the aforementioned discussions, we conclude the following: 
(1) each feature has advantages in certain scenes; 
(2) motion speed strongly correlates with the effectiveness of motion and appearance features. 
Inspired by this, we propose a parallel association paradigm to jointly utilize both features, and present the Two rOund Parallel matchIng meChanism (TOPIC) to implement it. 
As shown in Figure~\ref{fig:pipeline}(c), the TOPIC simultaneously uses motion and appearance features as assignment metrics, resembles taking the ``union'' of the matching results (reducing FN).
Additionally, TOPIC adaptively selects preferable matches based on motion level rather than filtering conflicting ones (reducing FN).
Except for association, detection and representation also impact tracking. Our proposed TOPICTrack adopts state-of-the-art YOLOX~\cite{ge2021yolox} detector, the motion model from OC-SORT~\cite{cao2023observation}, and the appearance model from FastReID~\cite{he2020fastreid}. Moreover, we propose an Attention-based Appearance Reconstruction Module (AARM) to enhance appearance representations.
Specifically, the AARM could improve the distinction among different objects' representations and enhance the similarity of representations for the same object across frames.

In this paper, we make contributions to the field of MOT by coping with the challenges posed by complex motion and diverse scenes through two key aspects: data construction and algorithm optimization. The main contributions of our work can be summarized as follows:
\begin{itemize}
\item We provide a dataset named BEE24, which highlights complex motion patterns, serving as a challenging benchmark for advancing general MOT algorithms research.
\item We propose a novel parallel association paradigm and design the TOPIC to implement it. The TOPIC utilizes motion and appearance features as association metrics in parallel and adaptively selects one of them according to the motion level to resolve conflicting matches. Additionally, the AARM is proposed to enhance trackers' ability to distinguish objects. 
\item 
Extensive experiments show the effectiveness and advantages of our proposed method on complex motions and diverse scenes. Our approach attains state-of-the-art results on five datasets in most metrics including HOTA, MOTA, and IDF1.
Furthermore, we demonstrate that our novel parallel association paradigm outperforms existing paradigms under fair comparisons, reducing false negatives by 6\% to 81\% on the five datasets versus the baseline. 
\end{itemize}

\section{Related Work}
\label{sec:formatting}
\subsection{Properties of MOT Datasets}
Illustrated in Figure~\ref{fig:data_motion}, current MOT datasets encompass diverse object categories and scenarios, including pedestrians~\cite{pets2009, chavdarova2018wildtrack, xu2018youtube, leal2015motchallenge, milan2016mot16, dendorfer2020mot20}, vehicles~\cite{geiger2012we, sun2020scalability}, group dances~\cite{sun2022dancetrack}, and even ants~\cite{wu2022dataset}.
Upon observing these datasets, it becomes evident that occlusion and highly similar appearances are mainly properties.
Occlusion, a commonly encountered property, introduces a considerable challenge in representing object appearance features, potentially rendering appearance features ineffective in extreme cases~\cite{liu2019model, liu2021end}.
On the other hand, highly similar appearances will reduce the visual distinction among different objects and pose a challenge for appearance-based trackers~\cite{henschel2020accurate, gao2022object}.
Moreover, through quantitative analysis, we conclude that existing datasets lack attention to complex motion patterns, with objects exhibiting simple motion patterns.
Specifically, the motion patterns of different objects are similar, and the motion intensity of the individual objects is low, showing small variations in successive frames.

However, more complex motion patterns are prevalent in life and nature, such as the phenomena like the activities of bee colonies around a hive~\cite{srinivasan1999motion}, as depicted in the top-left corner of Figure~\ref{fig:data_motion}.
To address the limitations of MOT datasets and explore the adaptability of trackers in coping with more complex scenarios, we provide a dataset focusing on bee colony activity, named BEE24. This dataset highlights the property of complex motion patterns, while also including occlusion and highly similar appearances.

\subsection{Appearance Feature-based Association}
Benefiting from the development of re-identification (re-ID)~\cite{shen2017learning, feng2019learning, tang2020cgan, lin2020multi, feng2021complementary, zhang2021attend, liu2021end, bai2021hierarchical,sun2021unsupervised},
most tracking algorithms rely mainly on appearance features for data association, regardless of the tracking-by-detection (TBD)~\cite{bewley2016simple, wojke2017simple, tang2017multiple, porzi2020learning, wang2021different} or joint detection and tracking (JDT) paradigms~\cite{wang2020towards, zhou2020tracking,ren2020tracking, peng2020chained,pang2021quasi,wu2021track, wan2021tracking, gurkan2021tdiot, zhang2021fairmot,meinhardt2022trackformer}.
Although the re-ID model relies on a large number of tracking annotations~\cite{karthik2020simple}, its ability to solve the task of object re-ID after long-term occlusion is important to this area of multi-object tracking.
The TBD paradigm treats detection and tracking as two independent tasks, e.g., the classical DeepSORT~\cite{wojke2017simple} uses a detector to obtain the location and size of objects and then builds a network to extract appearance embeddings.
The JDT paradigm, which has gained popularity in recent years, aims to combine detection and appearance feature extraction tasks, such as Mots~\cite{voigtlaender2019mots} and JDE~\cite{wang2020towards}.
They use a shared backbone network for end-to-end detection and appearance feature extraction. 
However, the performance of this paradigm is significantly degraded compared to the TBD paradigm, which is thought to be since training the re-ID model is complex~\cite{karthik2020simple, zhang2021fairmot}. FairMOT~\cite{zhang2021fairmot}, CTracker~\cite{peng2020chained}, and TraDeS~\cite{wu2021track} explore the compatible ways of re-ID and detection models, and achieve better tracking results than the TBD paradigm.
Recently, transformer shines in computing vision, and some works attempt to introduce attention mechanisms to learn appearance features~\cite{sun2020transtrack,meinhardt2022trackformer}.
Some work exploits scene information to improve the robustness of appearance models~\cite{li2023inference, fischer2023qdtrack}.
In the phase of data association, the aforementioned trackers generate appearance embeddings for current detections and historical trajectories. These embeddings are then used to compute their similarity for matching identities.

Despite the strides made by deep learning in enhancing the representation of appearances~\cite{hsu2021multi}, the reliability of appearance features diminishes in scenes with occlusions or highly similar appearances~\cite{sun2022dancetrack}. 
To this end, we introduce an attention-based appearance reconstruction module to augment appearance representation capability.

\subsection{Motion Feature-based Association}

Motion features are an effective cue used for data association.
Classical modeling techniques for motion features include Particle Filter~\cite{carpenter1999improved}, and Kalman Filter (KF)~\cite{welch1995introduction}.
These techniques operate on the assumption of linear object movement, utilizing past motion states to estimate present ones.
Due to the computational efficiency, the majority of prevalent trackers lean towards using the KF for extracting motion features, such as SORT~\cite{bewley2016simple} and DeepSORT~\cite{wojke2017simple}.
In the past, motion features were often employed as the auxiliary cue.
Take DeepSORT~\cite{wojke2017simple} for instance, which utilizes the KF to filter objects with abrupt shifts in motion patterns.
Recent studies obtain remarkable tracking results by making some improvements to the KF, which shows the importance of motion feature, including ByteTrack~\cite{zhang2022bytetrack}, FastTrack~\cite{liu2023fasttrack}, Decode-MOT~\cite{lee2023decode}, BPMTrack~\cite{gao2024bpmtrack} and OC-SORT~\cite{cao2023observation}.

However, the linear assumptions of such algorithms regarding motion patterns make tracking complex motion scenarios challenging.
An instance is the activities of bee colonies around their hive, as depicted in the first example of Figure~\ref{fig:data_motion}.
This highlights the necessity for existing trackers to enhance their performance by considering the combining of multiple features (e.g., appearance and motion).
As such, we argue that analyzing the suitability conditions of different features and then designing association paradigms to leverage their strengths effectively offers a promising way for MOT algorithms toward complex and diverse scenes.

\section{A New MOT Dataset}
\subsection{Dataset Construction}
\label{sec: Dataset Construction}

Currently, the main challenges of MOT datasets include occlusion (e.g., MOT17~\cite{milan2016mot16}, MOT20~\cite{dendorfer2020mot20}) and highly similar appearances (e.g., DanceTrack~\cite{sun2022dancetrack}, GMOT-40~\cite{bai2021gmot}, ANTS~\cite{wu2022dataset}).
Such challenges will degrade the performance of MOT algorithms that rely only on the appearance feature, thus adding other information (e.g., motion feature) would be beneficial~\cite{sun2022dancetrack}.
Existing motion models~\cite{carpenter1999improved,welch1995introduction,zhang2022bytetrack} can work well for low-speed and linear scenes.
However, it is challenging to cope with complex scenes, such as a bee switching from stationary to flying in a flash.

In this paper, we construct a bee dataset named BEE24. This dataset aims to highlight complex motion patterns while inheriting the properties of occlusion and highly similar appearance.
The property involves two aspects, differences in the motion patterns among different objects at the same frame and the variation for the individual object across frames.

To this end, we deployed visual monitor systems on five beehives to acquire videos of bee colony activities (refer to Supplementary Section~VII for more details). 
In order to capture a diverse range of bee colony activities, we acquired 36 videos (with a frame rate of 25 FPS and a resolution of 950$\times$590) during various periods on several sunny days.
For annotation, we use DarkLabel 2.1~\footnote{https://github.com/darkpgmr/DarkLabel}, a free and public MOT labeling software to annotate the dataset.
We employed five annotators to check and correct all labels to improve the quality of the dataset. 
The records format is the same as MOT17~\cite{milan2016mot16}.
The complete dataset is available at \url{https://drive.google.com/file/d/1KvqQTZWWhyDsQuEIqQY3XLM_8QlYjDqi/view?usp=sharing}.

\subsection{Dataset Statistics and Analysis}

\begin{table}[h]
\caption{Comparison of dataset meta-information between
BEE24 and popular MOT benchmarks.
$\downarrow$ means the smaller the better, others are the bigger the better.
\textbf{Bolding} and \underline{underline} denote the best and second-best results in each row, respectively.
Note that the average object area for DanceTrack is statistically based on the training set, as annotations for the test set are not publicly available.
}
\label{tab:comparsion_dataset}
\centering
\scalebox{0.77}{\begin{tabular}{l|llllll}
\toprule
Dataset                             & MOT17 & MOT20 & TAO&DanceTrack & GMOT-40 & BEE24 \\ \midrule
Videos                              & 42    & 8  &  \textbf{3,825} & \underline{100}        & 40      & 36    \\
Avg. len. (s)                         & 33    & \textbf{66}  &38  & \underline{52}         & 9       & 26    \\
Total len. (s)                       & 1,389  & 535  &\textbf{148,235} & \underline{5,292}       & 385     & 942   \\
Max. len. (s)                         & 62    & \underline{132} &42  & 120        & 29      & \textbf{200}   \\
Avg. tracks                          & 113   & \textbf{479} &4  & 9          & 49      & \underline{126}   \\
Total tracks                        & \underline{4,743}  & 3,833 &\textbf{16,104} & 990        & 1,944    & 4,559  \\
Max. tracks                          & 222   & \underline{1,211} &7 & 40         & 128     & \textbf{1,961}  \\
Avg. labels (10$^4$)   & \underline{3}     & \textbf{26}  &1  & 1          & 1       & 1     \\
Total labels (10$^4$) & \underline{118}   & \textbf{210}  &33 & 87         & 26      & 45    \\
Max. labels (10$^4$)   & 11    & \textbf{75} &8   & 2          & 2       & \underline{20}    \\
Avg. object area ($\downarrow$)                                 & 32,383    & 13,631 &31,023   & 68,078         & \underline{12,892}      & \textbf{3,099}    \\
FPS                                 & 24    & \underline{25}  & \textbf{30}  & 20         & \underline{25}      & \underline{25}    \\ \bottomrule
\end{tabular}}
\vspace{-0.5cm}
\end{table}

Some basic information about BEE24 can be found in Table~\ref{tab:comparsion_dataset}. 
BEE24 includes a total of 36 videos and 446,908 annotations, which are on par with those of popular MOT datasets, ensuring its scale is substantial enough for effective evaluation 
(see Supplementary Table~VIII, Figures~11-13 for more information on each video).
BEE24 has a much larger maximum duration (i.e., 200 s) and number of tracks for a single video than several common MOT datasets. For example, the maximum duration and tracks are an order of magnitude larger than those in GMOT-40.
The average occupied pixel area for a single object in BEE24 is one-fourth of that in the second-ranked dataset, GMOT-40, highlighting the challenge of detecting and tracking smaller objects.
Furthermore, the maximum number of annotations for a single video in BEE24 far exceeds those of other datasets, except for MOT20. MOT20 focuses on crowded scenes and therefore has the highest number of annotations. However, the objects' appearances are easily identifiable, and their slow motion tends to be linear. 
TAO contains thousands of categories of scenes and has the longest total duration; however, the maximum video length and tracks of TAO are relatively small.
Although DanceTrack achieves the highest total duration by increasing the number of videos, we argue that longer single video lengths pose a greater challenge, as it requires the ability of trackers to keep long-term track.

As discussed in Section~\ref{sec: Dataset Construction} for the property of complex motion patterns, we propose two metrics to analyze and compare the complexity of different MOT datasets.
First of all, in order to standardize the motion intensity $S$ of objects with different sizes, we introduce the following computation equation:
\begin{equation}
S = \sqrt{\left(\frac{x_2 - x_1}{w}\right)^2 + \left(\frac{y_2 - y_1}{h}\right)^2}.
\end{equation}
where $x_1$ and $x_2$ represent the horizontal displacements of the object across consecutive frames, $y_1$ and $y_2$ represent the vertical displacements, $w$ and $h$ denote the width and height of the object, respectively.

\noindent \textbf{\textit{Max-Min Speed Among Objects (MMSAO)}}: A metric to measure the difference of motion patterns among objects within the scene. A lower value implies greater similarity in motion patterns among different objects and vice versa. Given a dataset $D$, $\text{MMSAO}$ is formulated as follows:
\begin{equation}
\text{MMSAO} = \frac{1}{\sum_{v=1}^V T_{v}} \sum_{v=1}^V \sum_{t=1}^{T_{v}} \left( \max_{i\in \left[1:N_{v}^t\right]} S_{v,i}^{t} -  \min_{i\in \left[1:N_{v}^t\right]} S_{v,i}^{t} \right).
\end{equation}
where $V$ denotes the number of videos,
$T_{v}$ denotes the number of frames of the $v$-th video, 
$S_{v,i}^{t}$ denote the motion intensity of $i$-th object in the $t$-th frame of the $v$-th video,
$N_{v}^t$ represents the number of objects in the $t$-th frame of the $v$-th video, 
$\left[1:N_{v}^t\right]$ denotes the set ranging from 1 to $N_{v}^t$.

\noindent \textbf{\textit{Max-Min Speed of Object ($\text{MMSO}$)}}: A measure of the degree of motion pattern variation for an individual object.
A lower value indicates a more consistent speed for a single object across frames, and vice versa.
Given a dataset $D$, $\text{MMSO}$ is defined as follows:
\begin{equation}
\text{MMSO} = \frac{1}{\sum_{v=1}^V M_{v}} \sum_{v=1}^V \sum_{i=1}^{M_{v}} \left( \max_{t\in \left[1:{T_v}\right]} S_{v,i}^{t} -  \min_{t\in \left[1:{T_v}\right]} S_{v,i}^{t} \right).
\end{equation}
where $M_{v}$ denotes the total number of objects in the $v$-th video, 
other notations are the same as defined in $\text{MMSAO}$.

The metrics MMSAO and MMSO build upon the motion intensity $S$ to provide a comprehensive analysis of the complexity of motion patterns across different datasets. While $S$ measures the basic motion intensity standardized by object size, MMSAO and MMSO extend this concept to quantify the diversity and variability of motion. MMSAO captures the range of motion intensities across different objects in the same frame, which is indicative of the complexity and interaction among multiple objects. MMSO, on the other hand, captures the temporal variability of an object's speed, which reflects the consistency or unpredictability of its motion pattern. These metrics provide a scientific basis for analyzing the complexity of motion patterns.

\begin{figure}
\setlength{\abovecaptionskip}{1pt}
\setlength{\belowcaptionskip}{1pt}
\centering
\includegraphics[width=0.45\textwidth]{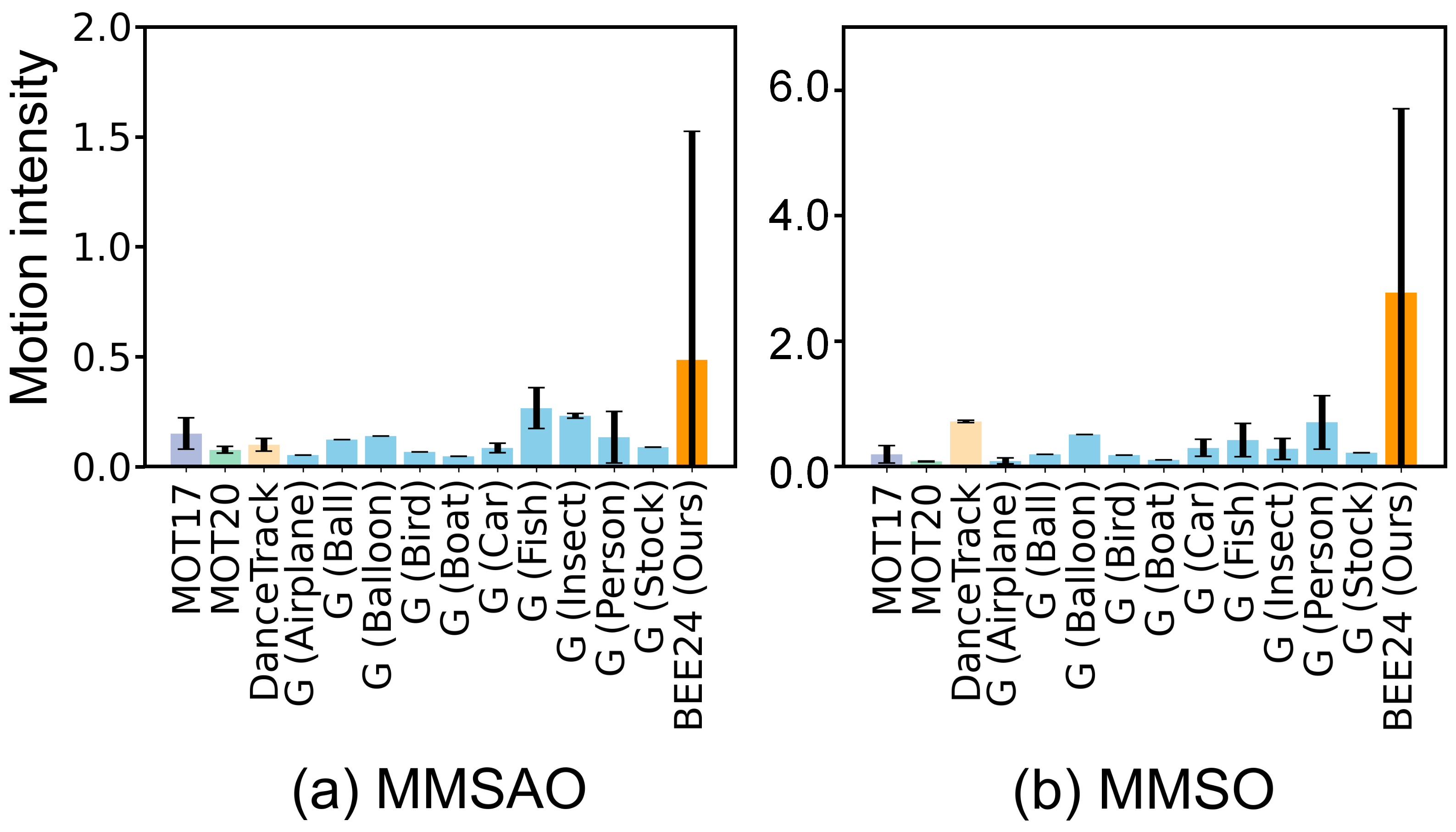}
\vspace{-0.5cm}
\hspace{0.5cm}
\caption{Comparison of motion pattern complexity between BEE24 and four popular datasets. 
(a) the diversity of motion patterns among objects;
(b) the variability of motion patterns of a single object across frames.
``G'' in xticks stands for GMOT-40.
}
\label{fig:MMSAO and MMSO}
\end{figure}

According to Figure~\ref{fig:MMSAO and MMSO}, we observe that  $\text{MMSAO}$ and  $\text{MMSO}$ metrics in BEE24 far exceed that of other datasets, which demonstrates that BEE24 highlights the property of complex motion patterns.
Although both GMOT-40 and BEE24 are natural scenes, the objects in GMOT-40 exhibit collective behavior, resulting in motion patterns that are often orderly and coordinated. In contrast, BEE24 features disordered and chaotic motion (see Supplementary Figure~16), with many individuals moving at high speeds (see Supplementary Figure~15). This further underscores the complex motion patterns in BEE24, presenting a new challenge for trackers.

Evaluating trackers on BEE24 can significantly enhance the performance of MOT algorithms in diverse real-world scenarios, such as the dynamic formations of dancers, drone swarms, and schools of fish. 
Additionally, the severe occlusions in BEE24 (Figure~\ref{fig:data_motion}) provide valuable insights for tracking tasks involving pedestrians, vehicles, drones, and other occlusion-prone objects.
Furthermore, the high similarity in the appearance of bees (Figure~\ref{fig:data_motion}) mirrors the challenges encountered when tracking animals and insects in nature, such as fish and ants, as well as products in industry, such as drones, thus increasing the generalizability of the tracker across different object classes.
Moreover, bees are among the smallest objects in popular existing datasets (Table~\ref{tab:comparsion_dataset}), making the development of algorithms capable of accurately tracking bees highly transferable to small object tracking tasks, such as monitoring insects or managing drone swarms captured from aerial perspectives.

To summarize, BEE24 challenges models to track multiple similar-appearing small objects with complex motions over long periods. It can serve as a testbed for MOT research toward more complex and diverse real-world scenes, significantly contributing to tracking general objects, including pedestrians, vehicles, drones, animals, insects, etc.

\section{Methodology}

\subsection{Two-Round Parallel Matching Mechanism}

\begin{figure}
\setlength{\abovecaptionskip}{1pt}
\setlength{\belowcaptionskip}{1pt}
\centering
\includegraphics[width=0.46\textwidth]{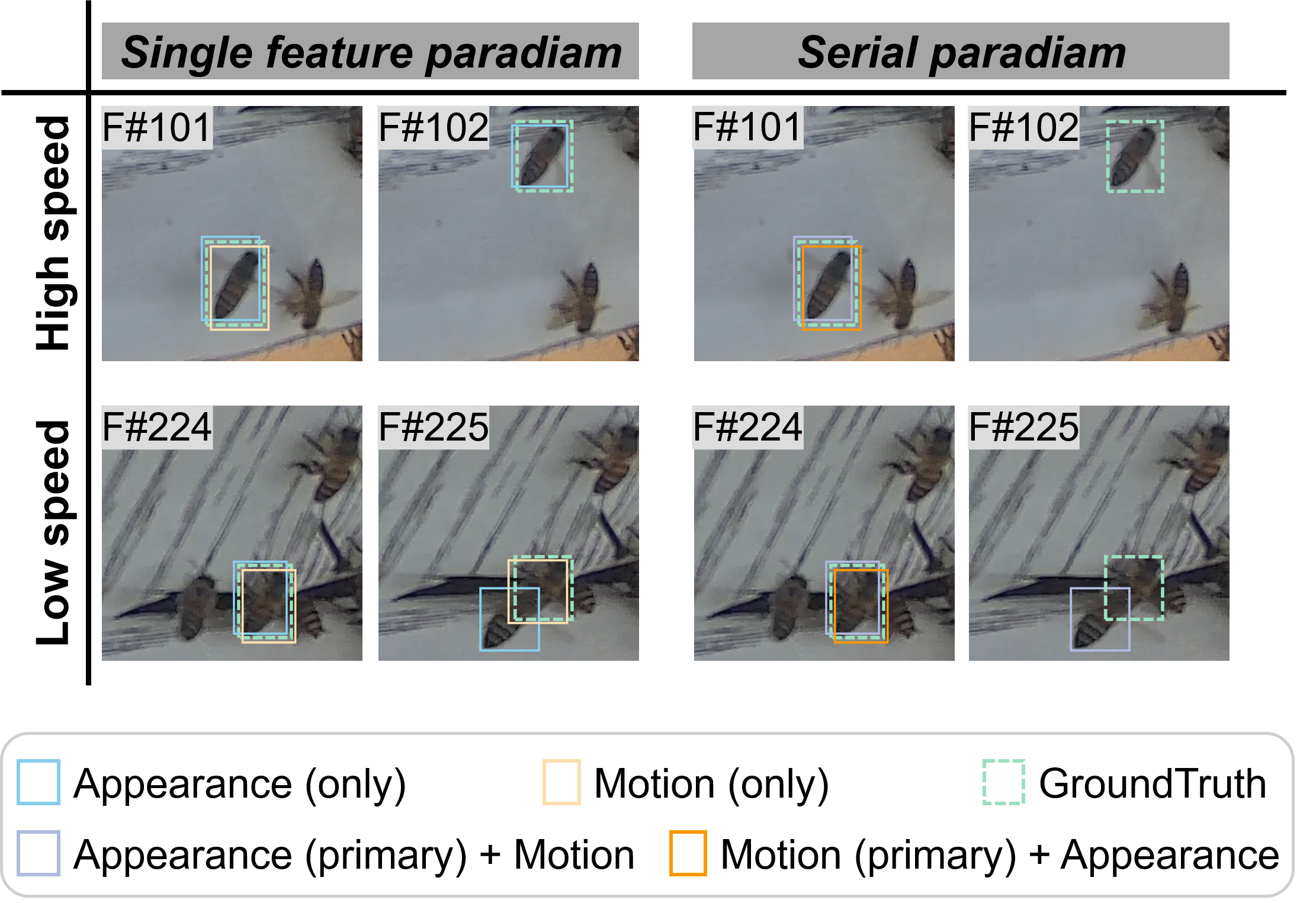}
\vspace{-0.2cm}
\caption{
Comparison of the performance of existing association paradigms on different scenes. 
The first row shows tracking a flying bee (high-speed);
the second row shows tracking an occluded bee (low-speed).
}
\label{fig:case_SOTA_failed}
\end{figure}

Existing identity association methods utilized by trackers can be categorized into two association paradigms, including the single-feature paradigm and the serial paradigm as mentioned in Figure~\ref{fig:pipeline}.
The single-feature paradigm either uses appearance features~\cite{fischer2023qdtrack} or motion features~\cite{zhang2022bytetrack, cao2023observation} as the assignment metric.
The serial paradigm, first manually selects a kind of feature as the filter to narrow the scope of association and then uses another feature as the primary assignment metric to finish the matching task~\cite{wojke2017simple, wang2020towards, peng2020chained, zhang2021fairmot, wu2021track}, avoiding conflicts from matching results of two features.
The intuition is that the more features used, the better the tracking performance.
How to utilize the advantages of both features to improve tracking performance through data association is the focus of this paper.

In the following, we first analyze the performance of existing association paradigms in different scenes.
As shown in the first row of Figure~\ref{fig:case_SOTA_failed}, a high-speed flying bee causes the motion feature to fail, thereby the association paradigm with the motion feature will fail.
However, in this situation, the flying bee will keep its distance to avoid collisions, resulting in its appearance being more visible~\cite{richardson2017swarm}. 
Hence, the association approach based on the appearance feature can successfully match the bee.
On the other side, when a bee moves slowly, as seen in the second row of Figure~\ref{fig:case_SOTA_failed}, the main challenges are occlusion and high appearance similarity, causing appearance features to be less reliable.
Therefore, applying the appearance feature for the association will yield an ID Switch.
In fact, in this case, the motion pattern of the bee tends to be linear, thus using a linear assumptions-based motion model for association could keep track of the bee.
Based on the above discussion, we conclude that:
(1) existing association paradigms cannot fully utilize advantages of different features according to different scenes;
(2) there is a strong correlation between motion speed and the effectiveness of motion and appearance features.

Empirical observations and a lot of practice have shown that motion models based on linear assumptions (e.g., KF) are suitable for the prediction of slow or linearly moving objects. However, when the object moves fast and nonlinearly, the uncertainty increases and these assumptions no longer hold, leading to increasing prediction errors~\cite{bar1985tracking, bewley2016simple, wojke2017simple, zhang2022bytetrack, cao2023observation}. Appearance models represent the object by extracting its appearance features, which requires the object's appearance to be as visually complete as possible. Particularly in high-density scenes, when the objects move slowly, it is easy to cause occlusion of each other, thus invalidating the appearance features~\cite{he2018deep}. In contrast, when objects move fast, occlusion is usually reduced, making it easier to recognize objects~\cite{perez2014idtracker}.

Based on the empirical observations of appearance and motion feature reliability, we propose a novel parallel association paradigm that uses motion and appearance features as assignment metrics in parallel, as shown in Figure~\ref{fig:pipeline}.
To implement this paradigm and to resolve match conflicts that may arise, we propose a \textbf{T}wo r\textbf{O}und \textbf{P}arallel match\textbf{I}ng me\textbf{C}hanism (TOPIC).

The objective of TOPIC is to maximize the total a posteriori probability of associations between trajectory sets $\{ \tau_i \}_n$ and detection sets $\{ \pi_j \}_m$ as follows:
\begin{equation}
    \max_M \sum_{(i,j) \in M} p(\pi_j | \tau_i)
\end{equation}
here, \( p(\pi_j | \tau_i) \) represents the probability that trajectory \(\tau_i\) and detection \(\pi_j\) are a correct match, which can be calculated based on appearance and motion features. Appearance feature-based posterior probabilities $p(\pi_j | \tau_i, \text{appearance})$ are measured by cosine distance~\cite{wojke2017simple}, and motion feature-based posterior probabilities $p(\pi_j | \tau_i, \text{motion})$ are measured by intersection over union (IOU) distance~\cite{cao2023observation}. 
We transform the posterior probabilities into association costs:
\begin{equation} 
\begin{aligned} 
A_{\text{cost}}(i, j) &= 1 - p_{\text{a}}(\tau_i | \pi_j), \\
\ M_{\text{cost}}(i, j) &= 1 - p_{\text{m}}(\tau_i | \pi_j). 
\end{aligned} 
\end{equation}
where $A_{cost}$ and $M_{cost}$ denote the cost matrices of appearance and motion features, respectively.
The matching probabilities between track $\tau_i$ and detection $\pi_j$ based on appearance and motion features are denoted by $p_{\text{a}}(\tau_i | \pi_j)$ and $ p_{\text{m}}(\tau_i | \pi_j)$, respectively.
Thereby, the optimization objective translates into minimizing the total matching costs of appearance and motion features, respectively:
\begin{equation}
\begin{alignedat}{2}
&\min_{M_a} &&\sum_{(i,j) \in M_a} A_{cost}(i, j) \\
&\min_{M_m} &&\sum_{(i,j) \in M_m} M_{cost}(i, j).
\end{alignedat}
\end{equation}

For the initialization of the algorithm, we calculate the appearance feature cost matrix $A_{cost}$ and the motion feature cost matrix $M_{cost}$ between tracklets and detections in parallel.
Then, the final matches $M$ is initialized to the empty set.
Next, the first round of matching is entered, where the TOPIC obtains appearance-based and motion-based matches $M_a$ and $M_m$ using the Hungarian algorithm~\cite{kuhn1955hungarian}, called pre-matching.
Within the results of pre-matching, the same matches $\hat{M}$ will be updated to the final matches $M$, while the conflicting matches $M_c$ enter the second round of matching.
In the second round, for matches with conflicts, the TOPIC adaptively selects matches derived from more reliable features according to the motion level, rather than filtering them out.

In this paper, we measure the motion level using IOU, and the formula is as follows:
\begin{equation}
\text{MotionLevel}_{i}^{t} = {1 - \text{IOU}(B_i^{t-k}, B_i^t)}
\end{equation}
where $\text{MotionLevel}_{i}^{t}$ indicates the motion level of tracklet $i$ at frame $t$, which takes values in $[0, 1]$. A larger value indicates a higher motion level and vice versa. 
Besides, $B_{i}^{t}$ denotes the bounding box of tracklet $i$ at $t$ frame, and $B_{i}^{t-k}$ represents the nearest observation box to frame $t$, applicable for both non-missing and missing detections.
Taking into account the uncertainty of object motion, we adopt a default assumption that when an object is initially tracked, its $\text{MotionLevel}=1$.

Moreover, we introduce a threshold of motion level $\alpha$, where $\alpha \in [0,1]$.
If $\text{MotionLevel}_{i}^{t} \geq \alpha$, we trust the results of appearance-based matching, otherwise we choose the results of motion-based matching. 
Note that the TOPIC degenerates to appearance-based and motion-based matching corresponding to $\alpha =0$ and $\alpha=1$, respectively.

After conflicts are resolved, we could obtain the final matches $M$, unmatched tracklets $un_T$, and unmatched detections $un_D$ for updating tracklets.
Algorithm~\ref{algo:Two-round matching mechanism} provides the pseudo-code of TOPIC.

\begin{algorithm}[!t]
\setlength{\abovecaptionskip}{1pt}
\setlength{\belowcaptionskip}{1pt}
\caption{Pseudo-code of the TOPIC}
\label{algo:Two-round matching mechanism}
\DontPrintSemicolon
\KwIn{Tracklets $\mathcal{T} = \{ \tau_i \}_n$; threshold of motion level $\alpha$; appearance-based cost matrix $A_{cost}$; motion-based cost matrix $M_{cost}$}
\KwOut{Matches $M$; unmatched tracklets $un_T$; unmatched detections $un_D$}
\textit{Initialization} $M \leftarrow \emptyset$

\tcp{First round: pre-matching based on the Hungarian algorithm}

$M_a, un_{T_a}, un_{D_a} \leftarrow $ \textit{Assignment} $A_{cost}$

$M_m, un_{T_m}, un_{D_m} \leftarrow $ \textit{Assignment} $M_{cost}$

$un_{T} \leftarrow $ \textit{Merge unmatched tracklets} $un_{T_a}, un_{T_m}$

$un_{D} \leftarrow $ \textit{Merge unmatched detections} $un_{D_a}, un_{D_m}$

\textit{Divide same matches} $\hat{M}$ \textit{and conflicted matches} $M_c$

$M \leftarrow M \cup \hat{M}$

\tcp{Second round: solving matching conflicts based on motion level}

\While{$M_c \neq \emptyset$}{
    Obtain tracklet indice $i$ from $M_{c}\left[0\right]$
    \If{the \text{MotionLevel} of $\tau_i$ is not lower than $\alpha$ }
    {
    Update $M$ by appearance-based matching
    }
    \Else{Update $M$ by motion-based matching}
    Update $M_c$
    }
\Return $M, un_T, un_D$
\end{algorithm}

\subsection{Attention-based Appearance Reconstruction Module}

\begin{figure*}[t!]
\setlength{\abovecaptionskip}{1pt}
\setlength{\belowcaptionskip}{1pt}
\centering
\includegraphics[width=0.58\textwidth]{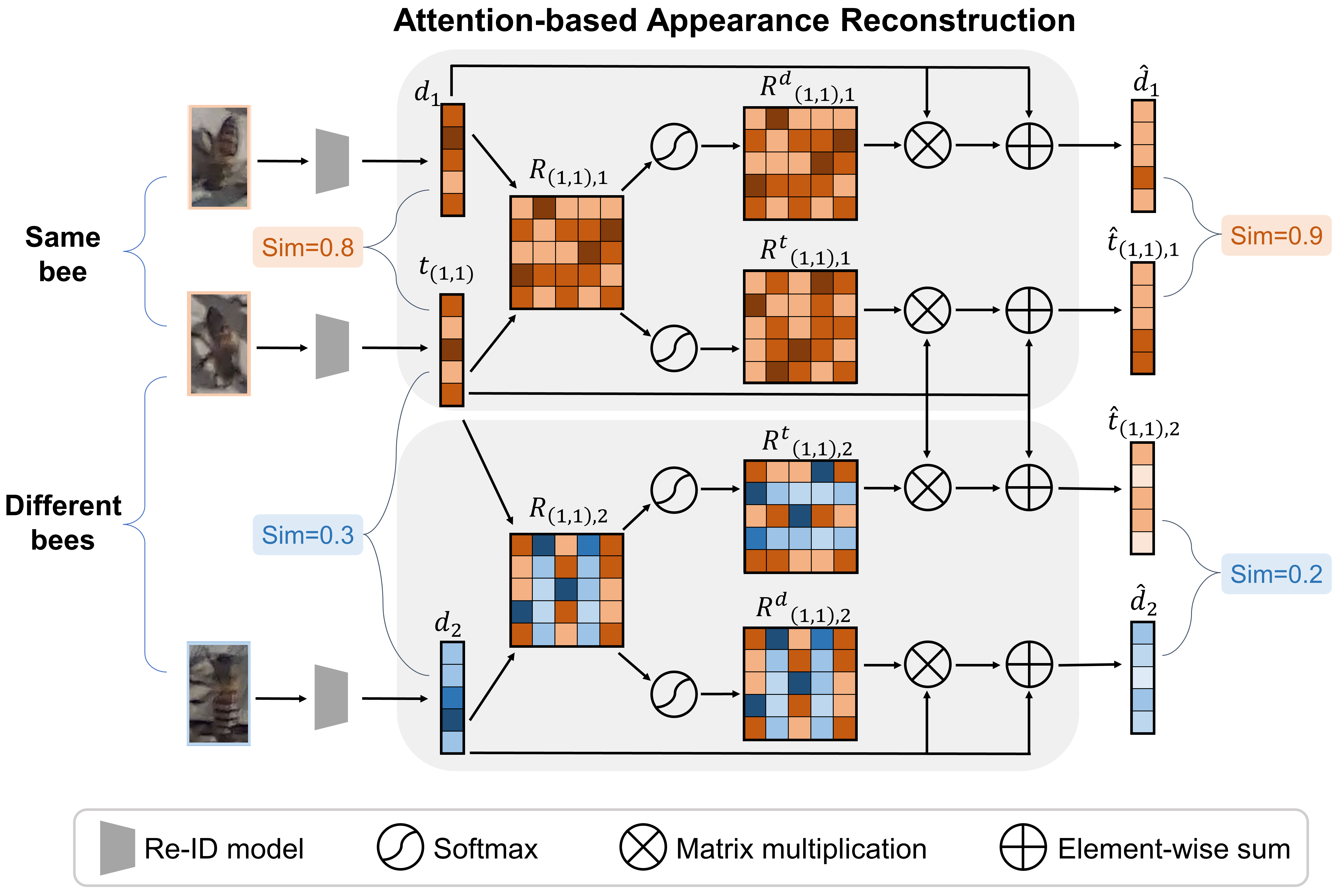}
\vspace{0.1cm}
\caption{Overview of the AARM.
Taking the similarity metric of the same bee as an example, we first use the history trajectory of the bee and the current detection's appearance embeddings $t_{(1,1)}$ and $d_1$ to compute the attention map $R_{(1,1),1}$.
Next, by softmax operation on the attention map to get $R^d_{(1,1),1}$, and then after transposition to get $R^t_{(1,1),1}$, thus obtaining two cross-attention maps.
Afterward, appearance embeddings $t_{(1,1)}$ and $d_1$ are reconstructed via the residual attention mechanism.
After reconstruction, the similarity score of appearance embeddings of the same bee is increased, e.g., from 0.8 to 0.9.
And vice versa for different bees.
}
\label{fig:reappearance-module}
\end{figure*}

Since the appearance model impacts the tracking performance, this sub-section aims to introduce a novel appearance feature extraction method to enhance the appearance representation capability.
Inspired by Fisher discrimination criterion~\cite{fisher1936use}, which maximizes the inter-class distance and minimizes the intra-class distance, we propose an attention-based appearance reconstruction module (AARM) to enhance the distinguishing ability for object identity. 
As shown in Figure~\ref{fig:reappearance-module}, the AARM could improve the distinction among different objects' representations and enhance the similarity of representations for the same object across frames.

Let $\{ \{ t_{i,k}\}_{k=1}^{L_i} \}_{i=1}^{n}$ denote the appearance embedding galleries of $n$ previous tracklets, where $t_{i,k} \in \mathbb{R}^{dim}$ indicates the $k$-th embedding of tracklet $i$, $dim$ indicates the embedding dimension, and $L_i$ denotes the number of embeddings of tracklet $i$. 
Similarly, $\{d_j \in \mathbb{R}^{dim}\}_{j=1}^m$ denotes appearance embeddings of the current $m$ detections.

Firstly, we use the cosine distance for the $k$-th embedding of tracklet $i$ and the embedding of detection $j$ to compute an attention map $R_{(i,k),j} \in \mathbb{R}^{dim \times dim}$ as:
\begin{equation}
R_{(i,k),j}=\left(\frac{t_{i,k}}{\left\|t_{i,k}\right\|_2}\right)^T\left(\frac{d_j}{\left\|d_j\right\|_2}\right).
\end{equation}
Next, we utilize matrix transposition and the \textit{Softmax} function to obtain two cross-attention matrixes:
\begin{equation}
\begin{aligned}
R_{(i,k),j}^d &= \textit{Softmax}(R_{(i,k),j}) \\
R_{(i,k),j}^t &= \textit{Softmax}(R_{(i,k),j})^T.
\end{aligned}
\end{equation}

Then, we use a residual attention mechanism to obtain two reconstructed appearance embeddings $\hat{d}_j$ and $\hat{t}_{(i,k),j}$:
\begin{equation}
\begin{aligned}
\hat{d}_{j} &= (R_{(i,k),j}^d+I)d_{j} \\
\hat{t}_{(i,k),j} &= (R_{(i,k),j}^t+I)t_{i,k}.
\end{aligned}
\end{equation}

Further, we calculate the appearance similarity of the reconstructed embeddings of tracklet $i$ and detection $j$ as follows:
\begin{equation}
A_{sim_{i,j}} = max\{{\hat{d}_j}^{T}\hat{t}_{\left(i,k\right),j}~| k \in \left[1, L_i\right] \}.
\end{equation}

Finally, we can obtain the appearance-based cost matrix $A_{cost} = (1-A_{sim_{i,j}})_{n \times m}$. It is worth mentioning that the appearance reconstruction module requires no training and can be plug-and-play to other trackers.

We analyze the principles underlying the ability of AARM to be effective.
Each element value in the cross-attention map represents the similarity between the tracklet and the detection features. The closer to 1 indicates that the features in that part are more similar, and vice versa. The same object in different frames gets more attention due to more similar parts, while vice versa for different objects in the same frame. Thus, the reconstructed appearance features enhance the ability to distinguish object identity.

\section{Experiments}
\subsection{Experiment Setup}
\label{sec:Experiment Setup}

\subsubsection{Dataset settings}
For a fair comparison on MOT17~\cite{milan2016mot16}, MOT20~\cite{dendorfer2020mot20} and DanceTrack~\cite{sun2022dancetrack}, our method follows the same dataset split as ByteTrack~\cite{zhang2022bytetrack}.
For GMOT-40~\cite{bai2021gmot}, we take the first three sequences of each class as the training set and the remaining one sequence as the test set.
For BEE24, we randomly select the 31 most representative videos as the training set, and the remaining 5 videos are used as the test set.

\subsubsection{Implementation details}
In this paper, we utilize YOLOX~\cite{ge2021yolox} as the detector, OC-SORT~\cite{cao2023observation} as the motion model, and FastReID~\cite{he2020fastreid} as the re-ID model in our proposed method.
For the ablation study, we ensure that the parameter settings of each method align with those specified in the original papers or the official open-source code base. 
For our proposed TOPICTrack, only the re-ID model needs to be trained, where we adopt the AGW model from FastReID with the Adam optimizer~\cite{kingma2014adam} for 120 epochs with a start learning rate of 3.5e-4. The learning rate decays to 5e-4. The input size is reshaped to 384$\times$384. The batch size is set to 64.
The training step takes about 3 hours on 2 NVIDIA RTX 3090 GPUs.

\subsubsection{Metrics}
We consider HOTA~\cite{luiten2021hota}, CLEAR metrics~\cite{bernardin2008evaluating}, IDF1 Score (IDF1)~\cite{ristani2016performance}, AssA~\cite{luiten2021hota} and AssR~\cite{luiten2021hota} as evaluation metrics. For the CLEAR metrics, we report Multiple-Object Tracking Accuracy (MOTA), the number of False Negatives (FN), False Positives (FP), Identity Switches (IDs), and Fragments (Frag). The MOTA accounts for all object tracking errors (including FP, FN, and IDs) made by the tracker over all frames.

\subsection{Ablation Studies}

\begin{table*}[!ht]
\centering
\caption{
Ablation studies of the association paradigm (TOPIC) and AARM on five datasets.
``A'' and ``M'' denote the appearance model from FastReID and the motion model from OC-SORT, respectively. 
``\textit{A-based}'' and ``\textit{M-based}'' denote using the appearance and motion features as assignment metrics, respectively.
``Location-based tracker'' denotes a baseline tracker that utilizes location-based distance (Euclidean distance) measures for association.
$\uparrow$ follows the metric indicating the larger value, the better performance, and vice versa.
\textcolor[RGB]{0,100,0}{\textbf{Green}} indicates the value improves more than 1.
}
\label{tab:ablation}
\setlength{\tabcolsep}{5pt}
\scriptsize
\scalebox{0.8}{
\begin{tabular}{cllllllllllll}
\toprule
\belowrulesepcolor{gray!70} 
\rowcolor{gray!70}
\multicolumn{7}{c|}{ \textit{\textbf{Single-Feature Association Paradiam}}}
& 
\multicolumn{6}{c}{\textit{\textbf{Serial Association Paradiam (Two Features)}}}
\\ 
\aboverulesepcolor{gray!70} 
\midrule

\belowrulesepcolor{gray!20} 
\rowcolor{gray!20}
\multicolumn{13}{c}{\textbf{MOT17 Validation Set}}                                        \\ 
\aboverulesepcolor{gray!20} 
\midrule

\multicolumn{1}{l}{}                                    & \multicolumn{1}{|l|}{Scheme}        & HOTA$\uparrow$ & MOTA$\uparrow$ & IDF1$\uparrow$ & FP$\downarrow$ & \multicolumn{1}{l|}{FN$\downarrow$} & \multicolumn{1}{l|}{Scheme}      & HOTA$\uparrow$ & MOTA$\uparrow$ & IDF1$\uparrow$ & FP$\downarrow$ & FN$\downarrow$ \\ 

\midrule

\multicolumn{1}{c}{\cellcolor{gray!70}}  
& \multicolumn{1}{|l|}{ByteTrack~\cite{zhang2022bytetrack}}     & 64.4           & 73.2           & 75.8           & 2,002          & \multicolumn{1}{l|}{11,975}         & \multicolumn{1}{l|}{-}           &         -       &        -        &  -              &      -          &      -          \\

\multicolumn{1}{c}{\cellcolor{gray!70}}
& \multicolumn{1}{|l|}{+A}            & 64.8 (+0.4)    & 73.4 (+0.2)    & 75.9 (+0.1)    & 2,057           & \multicolumn{1}{l|}{11,797}          & \multicolumn{1}{l|}{TraDeS~\cite{wu2021track}}      & 58.7           & 68.5           & 71.8           & 1,693          & 12,445         \\

\multicolumn{1}{c}{\cellcolor{gray!70}}  
& \multicolumn{1}{|l|}{+A+TOPIC}      & 66.8 (\textcolor[RGB]{0,100,0}{\textbf{+2.4}})    & 76.1 (\textcolor[RGB]{0,100,0}{\textbf{+2.9}})    & 77.3 (\textcolor[RGB]{0,100,0}{\textbf{+1.5}})    & 2,264           & \multicolumn{1}{l|}{9,923}           & \multicolumn{1}{l|}{+TOPIC}      & 60.1 (\textcolor[RGB]{0,100,0}{\textbf{+1.4}})    & 69.5 (\textcolor[RGB]{0,100,0}{\textbf{+1.0}})    & 72.8 (\textcolor[RGB]{0,100,0}{\textbf{+1.0}})    & 2,421           & 9,487           \\

\multicolumn{1}{c}{\cellcolor{gray!70}}
& \multicolumn{1}{|l|}{+A+AARM}       & 66.0 (\textcolor[RGB]{0,100,0}{\textbf{+1.6}})    & 75.7 (\textcolor[RGB]{0,100,0}{\textbf{+2.5}})    & 76.7 (+0.9)    & 2,355          & \multicolumn{1}{l|}{9,807}          & \multicolumn{1}{l|}{+AARM}       & 59.6 (+0.9)    & 69.4 (+0.9)    & 72.5 (+0.7)    & 2,476          & 9,437          \\
\multicolumn{1}{c}{\multirow{-5}{*}
{\cellcolor{gray!70} \rotatebox{90}{\textit{M-Based}}}}
& \multicolumn{1}{|l|}{+A+TOPIC+AARM} & 67.7 (\textcolor[RGB]{0,100,0}{\textbf{+3.3}})    & 77.8 (\textcolor[RGB]{0,100,0}{\textbf{+4.6}})    & 79.7 (\textcolor[RGB]{0,100,0}{\textbf{+3.9}})    & 2,651          & \multicolumn{1}{l|}{9,125}          & \multicolumn{1}{l|}{+TOPIC+AARM} & 60.4 (\textcolor[RGB]{0,100,0}{\textbf{+1.7}})    & 69.9 (\textcolor[RGB]{0,100,0}{\textbf{+1.4}})    & 73.8 (\textcolor[RGB]{0,100,0}{\textbf{+2.0}})     & 2,685          & 9,048          \\ \midrule

\multicolumn{1}{c}{\cellcolor{gray!70}}                                   & \multicolumn{1}{|l|}{Location-based tracker}  & 60.8 & 74.1 & 68.1 & 1,529 & \multicolumn{1}{l|}{11,764}          & \multicolumn{1}{l|}{-}           &      -          &       -         &            -    &       -         &   -             \\
\multicolumn{1}{c}{\cellcolor{gray!70}}                                   & \multicolumn{1}{|l|}{+A}            & 66.4 (+0.3)          & 74.7           & 77.8           & 1,290           & \multicolumn{1}{l|}{12,140} & \multicolumn{1}{l|}{-}           &      -          &       -         &            -    &       -         &   -             \\
\multicolumn{1}{c}{\cellcolor{gray!70}}                                   & \multicolumn{1}{|l|}{+M}            & 66.9 (+0.5)    & 74.9 (+0.2)    & 78.1 (+0.3)    & 1,302           & \multicolumn{1}{l|}{12,010}          & \multicolumn{1}{l|}{FairMOT~\cite{zhang2021fairmot}}     & 60.5           & 67.5           & 69.9           & 3,729          & 6,743          \\

\multicolumn{1}{c}{\cellcolor{gray!70}}                                   & \multicolumn{1}{|l|}{+M+A+TOPIC}      & 68.0 (\textcolor[RGB]{0,100,0}{\textbf{+2.6}})    & 79.7 (\textcolor[RGB]{0,100,0}{\textbf{+5.0}})    & 79.9 (\textcolor[RGB]{0,100,0}{\textbf{+2.1}})    & 2,128           & \multicolumn{1}{l|}{10,343}          & \multicolumn{1}{l|}{+TOPIC}      & 61.9 (\textcolor[RGB]{0,100,0}{\textbf{+1.4}})    & 68.4 (+0.9)    & 70.9 (\textcolor[RGB]{0,100,0}{\textbf{+1.0}})    & 3,802           & 5,734           \\

\multicolumn{1}{c}{\cellcolor{gray!70}}                                   & \multicolumn{1}{|l|}{+AARM}         & 68.7 (\textcolor[RGB]{0,100,0}{\textbf{+2.3}})    & 79.5 (\textcolor[RGB]{0,100,0}{\textbf{+4.8}})    & 79.4 (\textcolor[RGB]{0,100,0}{\textbf{+1.6}})    & 2,978           & \multicolumn{1}{l|}{7,687}           & \multicolumn{1}{l|}{+AARM}       & 61.6 (\textcolor[RGB]{0,100,0}{\textbf{+1.1}})    & 68.3 (+0.8)    & 70.4 (+0.5)    & 3,876          & 5,684          \\

\multicolumn{1}{c}{\multirow{-5}{*}{\cellcolor{gray!70} \rotatebox{90}{\textit{A-Based}}}}
& \multicolumn{1}{|l|}{+M+TOPIC+AARM} & 69.9 (\textcolor[RGB]{0,100,0}{\textbf{+3.5}})    & 79.8 (\textcolor[RGB]{0,100,0}{\textbf{+5.1}})    & 81.6 (\textcolor[RGB]{0,100,0}{\textbf{+3.8}})    & 3,065           & \multicolumn{1}{l|}{7,568}           & \multicolumn{1}{l|}{+TOPIC+AARM} & 62.3 (\textcolor[RGB]{0,100,0}{\textbf{+1.8}})     & 68.5 (\textcolor[RGB]{0,100,0}{\textbf{+1.0}})    & 72.5 (\textcolor[RGB]{0,100,0}{\textbf{+2.6}})    & 4,021          & 5,456          \\ \midrule

\belowrulesepcolor{gray!20} 
\rowcolor{gray!20}
\multicolumn{13}{c}{\textbf{MOT20 Validation Set}}         \\
\aboverulesepcolor{gray!20} 
\midrule

\multicolumn{1}{c}{\cellcolor{gray!70}}                                   & \multicolumn{1}{|l|}{ByteTrack~\cite{zhang2022bytetrack}}     & 54.7           & 68.2           & 70.2           & 24,767         & \multicolumn{1}{l|}{184,357}        & \multicolumn{1}{l|}{-}           &       -         &    -            &    -            &    -            &      -          \\

\multicolumn{1}{c}{\cellcolor{gray!70}}                                   & \multicolumn{1}{|l|}{+A}            & 54.9 (+0.2)    & 68.6 (+0.4)    & 70.5 (+0.3)    & 24,945          & \multicolumn{1}{l|}{182,453}         & \multicolumn{1}{l|}{TraDeS~\cite{wu2021track}}      & 46.6           & 65.3           & 67.5           & 31,765         & 186,456        \\

\multicolumn{1}{c}{\cellcolor{gray!70}}                                   & \multicolumn{1}{|l|}{+A+TOPIC}      & 56.3 (\textcolor[RGB]{0,100,0}{\textbf{+1.6}})    & 69.7 (\textcolor[RGB]{0,100,0}{\textbf{+1.5}})    & 71.7 (\textcolor[RGB]{0,100,0}{\textbf{+1.5}})    & 25,215          & \multicolumn{1}{l|}{161,978}         & \multicolumn{1}{l|}{+TOPIC}      & 47.8 (\textcolor[RGB]{0,100,0}{\textbf{+1.2}})    & 67.9 (\textcolor[RGB]{0,100,0}{\textbf{+2.6}})    & 67.8 (+0.3)    & 32,078          & 165,453         \\

\multicolumn{1}{c}{\cellcolor{gray!70}}                                   & \multicolumn{1}{|l|}{+A+AARM}       & 55.6 (+0.9)    & 69.1 (+0.9)    & 71.0 (+0.8)    & 25,606         & \multicolumn{1}{l|}{161,650}        & \multicolumn{1}{l|}{+AARM}       & 47.2 (+0.6)    & 67.7 (\textcolor[RGB]{0,100,0}{\textbf{+2.4}})    & 67.8 (+0.3)    & 32,342         & 165,323        \\
\multicolumn{1}{c}{\multirow{-5}{*}{\cellcolor{gray!70} \rotatebox{90}{\textit{M-Based}}}}          & \multicolumn{1}{|l|}{+A+TOPIC+AARM} & 56.9 (\textcolor[RGB]{0,100,0}{\textbf{+2.2}})    & 72.7 (\textcolor[RGB]{0,100,0}{\textbf{+4.5}})    & 72.4 (\textcolor[RGB]{0,100,0}{\textbf{+2.2}})    & 30,494         & \multicolumn{1}{l|}{136,443}        & \multicolumn{1}{l|}{+TOPIC+AARM} & 48.1 (\textcolor[RGB]{0,100,0}{\textbf{+1.5}})    & 68.6 (\textcolor[RGB]{0,100,0}{\textbf{+3.3}})    & 67.9 (+0.4)    & 34,856         & 156,742        \\ \midrule

\multicolumn{1}{c}{\cellcolor{gray!70}}                                   & \multicolumn{1}{|l|}{Location-based tracker}  & 43.1           & 63.3           & 49.7           & 19,852          & \multicolumn{1}{l|}{199,095}         & \multicolumn{1}{l|}{-}           &     -           &       -         &            -    &        -        &     -           \\
\multicolumn{1}{c}{\cellcolor{gray!70}}                                   & \multicolumn{1}{|l|}{+A}            & 56.5           & 70.0           & 71.3           & 27,815          & \multicolumn{1}{l|}{155,511}           &      -          &       -         &            -    &       -         &   -             \\

\multicolumn{1}{c}{\cellcolor{gray!70}}                                   & \multicolumn{1}{|l|}{+M}            & 56.9 (+0.4)    & 70.3 (+0.3)    & 71.6 (+0.3)    & 27,902          & \multicolumn{1}{l|}{155,356}         & \multicolumn{1}{l|}{FairMOT~\cite{zhang2021fairmot}}     & 40.2           & 60.2           & 60.8           & 35,684         & 194,346        \\

\multicolumn{1}{c}{\cellcolor{gray!70}}                                   & \multicolumn{1}{|l|}{+M+A+TOPIC}      & 57.3 (+0.8)    & 72.3 (\textcolor[RGB]{0,100,0}{\textbf{+2.3}})    & 72.0 (+0.7)    & 28,193          & \multicolumn{1}{l|}{136,799}         & \multicolumn{1}{l|}{+TOPIC}      & 41.2 (\textcolor[RGB]{0,100,0}{\textbf{+1.0}})    & 62.9 (\textcolor[RGB]{0,100,0}{\textbf{+2.7}})    & 60.9 (+0.1)    & 36,181          & 176,815         \\

\multicolumn{1}{c}{\cellcolor{gray!70}}                                   & \multicolumn{1}{|l|}{+AARM}         & 56.9 (+0.4)    & 71.3 (\textcolor[RGB]{0,100,0}{\textbf{+1.3}})    & 71.6 (+0.3)    & 28,402          & \multicolumn{1}{l|}{136,865}         & \multicolumn{1}{l|}{+AARM}       & 40.9 (+0.7)    & 62.7 (\textcolor[RGB]{0,100,0}{\textbf{+2.5}})    & 60.9 (+0.1)    & 36,543         & 176,543        \\
\multicolumn{1}{c}{\multirow{-5}{*}{\cellcolor{gray!70} \rotatebox{90}{\textit{A-Based}}}}          & \multicolumn{1}{|l|}{+M+TOPIC+AARM} & 57.6 (\textcolor[RGB]{0,100,0}{\textbf{+1.1}})    & 73.0 (\textcolor[RGB]{0,100,0}{\textbf{+3.0}})    & 72.3 (\textcolor[RGB]{0,100,0}{\textbf{+1.0}})    & 28,702          & \multicolumn{1}{l|}{135,881}         & \multicolumn{1}{l|}{+TOPIC+AARM} & 41.7 (\textcolor[RGB]{0,100,0}{\textbf{+1.5}})    & 63.5 (\textcolor[RGB]{0,100,0}{\textbf{+3.3}})    & 61.1 (+0.3)    & 37,764         & 172,681        \\ \midrule

\belowrulesepcolor{gray!20} 
\rowcolor{gray!20}
\multicolumn{13}{c}{\textbf{DanceTrack Validation Set}}    \\ 
\aboverulesepcolor{gray!20} 
\midrule

\multicolumn{1}{c}{\cellcolor{gray!70}}                                   & \multicolumn{1}{|l|}{ByteTrack~\cite{zhang2022bytetrack}}     & 34.5           & 78.5           & 30.0           & 2,657          & \multicolumn{1}{l|}{18,043}         & \multicolumn{1}{l|}{-}           &      -          &      -          &   -             &         -       &   -             \\

\multicolumn{1}{c}{\cellcolor{gray!70}}                                   & \multicolumn{1}{|l|}{+A}            & 34.6 (+0.1)    & 78.7 (+0.2)    & 30.4 (+0.4)    & 2,721           & \multicolumn{1}{l|}{17,867}          & \multicolumn{1}{l|}{TraDeS~\cite{wu2021track}}      & 41.4           & 80.3           & 40.2           & 4,067          & 6,674          \\

\multicolumn{1}{c}{\cellcolor{gray!70}}                                   & \multicolumn{1}{|l|}{+A+TOPIC}      & 35.9 (\textcolor[RGB]{0,100,0}{\textbf{+1.4}})    & 79.8 (\textcolor[RGB]{0,100,0}{\textbf{+1.3}})    & 31.1 (\textcolor[RGB]{0,100,0}{\textbf{+1.1}})    & 5,365           & \multicolumn{1}{l|}{14,102}          & \multicolumn{1}{l|}{+TOPIC}      & 43.0 (\textcolor[RGB]{0,100,0}{\textbf{+1.9}})    & 81.4 (\textcolor[RGB]{0,100,0}{\textbf{+1.1}})    & 41.1 (+0.9)    & 6,023           & 4,026           \\

\multicolumn{1}{c}{\cellcolor{gray!70}}                                   & \multicolumn{1}{|l|}{+A+AARM}       & 35.3 (+0.8)    & 79.4 (+0.9)    & 30.8 (+0.8)    & 5,587          & \multicolumn{1}{l|}{13,984}         & \multicolumn{1}{l|}{+AARM}       & 42.6 (\textcolor[RGB]{0,100,0}{\textbf{+1.2}})    & 81.2 (+0.9)    & 40.9 (+0.7)    & 6,456          & 3,658          \\
\multicolumn{1}{c}{\multirow{-5}{*}{\cellcolor{gray!70} \rotatebox{90}{\textit{M-Based}}}}          & \multicolumn{1}{|l|}{+A+TOPIC+AARM} & 36.6 (\textcolor[RGB]{0,100,0}{\textbf{+2.1}})    & 80.4 (\textcolor[RGB]{0,100,0}{\textbf{+1.9}})    & 31.3 (\textcolor[RGB]{0,100,0}{\textbf{+1.3}})    & 5,606          & \multicolumn{1}{l|}{12,478}         & \multicolumn{1}{l|}{+TOPIC+AARM} & 43.8 (\textcolor[RGB]{0,100,0}{\textbf{+2.4}})    & 81.9 (\textcolor[RGB]{0,100,0}{\textbf{+1.6}})    & 41.6 (\textcolor[RGB]{0,100,0}{\textbf{+1.4}})    & 6,574          & 2,541          \\ \midrule

\multicolumn{1}{c}{\cellcolor{gray!70}}                                   & \multicolumn{1}{|l|}{Location-based tracker}  & 40.6           & 84.8           & 34.1           & 5,526          & \multicolumn{1}{l|}{24,867}          & \multicolumn{1}{l|}{-}           &      -          &      -          &            -    &        -        &    -            \\
\multicolumn{1}{c}{\cellcolor{gray!70}}                                   & \multicolumn{1}{|l|}{+A}            & 52.4           & 87.3           & 52.1           & 12,155          & \multicolumn{1}{l|}{24,526} & \multicolumn{1}{l|}{-}           &      -          &       -         &            -    &       -         &   -             \\

\multicolumn{1}{c}{\cellcolor{gray!70}}                                   & \multicolumn{1}{|l|}{+M}            & 52.7 (+0.3)    & 87.5 (+0.2)    & 52.6 (+0.5)    & 12,203          & \multicolumn{1}{l|}{24,211}          & \multicolumn{1}{l|}{FairMOT~\cite{zhang2021fairmot}}     & 37.3           & 79.1           & 39.5           & 4,155          & 37,526         \\

\multicolumn{1}{c}{\cellcolor{gray!70}}                                   & \multicolumn{1}{|l|}{+M+A+TOPIC}      & 54.5 (\textcolor[RGB]{0,100,0}{\textbf{+2.1}})    & 88.4 (\textcolor[RGB]{0,100,0}{\textbf{+1.1}})    & 53.9 (\textcolor[RGB]{0,100,0}{\textbf{+1.8}})    & 10,159          & \multicolumn{1}{l|}{14,071}          & \multicolumn{1}{l|}{+TOPIC}      & 38.5 (\textcolor[RGB]{0,100,0}{\textbf{+1.2}})    & 80.6 (\textcolor[RGB]{0,100,0}{\textbf{+1.5}})    & 40.9 (\textcolor[RGB]{0,100,0}{\textbf{+1.4}})    & 4,801           & 33,261          \\

\multicolumn{1}{c}{\cellcolor{gray!70}}                                   & \multicolumn{1}{|l|}{+AARM}         & 54.1 (\textcolor[RGB]{0,100,0}{\textbf{+1.7}})    & 88.3 (\textcolor[RGB]{0,100,0}{\textbf{+1.0}})    & 52.5 (+0.4)    & 12,572          & \multicolumn{1}{l|}{10,950}          & \multicolumn{1}{l|}{+AARM}       & 38.2 (+0.9)    & 80.4 (\textcolor[RGB]{0,100,0}{\textbf{+1.3}})    & 40.8 (\textcolor[RGB]{0,100,0}{\textbf{+1.3}})    & 5,642          & 32,546         \\
\multicolumn{1}{c}{\multirow{-5}{*}{\cellcolor{gray!70} \rotatebox{90}{\textit{A-Based}}}}          & \multicolumn{1}{|l|}{+M+TOPIC+AARM} & 55.9 (\textcolor[RGB]{0,100,0}{\textbf{+3.5}})    & 89.3 (\textcolor[RGB]{0,100,0}{\textbf{+2.0}})    & 54.5 (\textcolor[RGB]{0,100,0}{\textbf{+2.4}})    & 12,816          & \multicolumn{1}{l|}{10,622}          & \multicolumn{1}{l|}{+TOPIC+AARM} & 38.9 (\textcolor[RGB]{0,100,0}{\textbf{+1.6}})    & 80.9 (\textcolor[RGB]{0,100,0}{\textbf{+1.8}})    & 41.2 (\textcolor[RGB]{0,100,0}{\textbf{+1.7}})    & 5,215          & 31,565         \\ \midrule

\belowrulesepcolor{gray!20} 
\rowcolor{gray!20}
\multicolumn{13}{c}{\textbf{GMOT-40 Test Set}}      \\
\aboverulesepcolor{gray!20} 
\midrule

\multicolumn{1}{c}{\cellcolor{gray!70}}                                   & \multicolumn{1}{|l|}{ByteTrack~\cite{zhang2022bytetrack}}     & 75.6           & 89.4           & 83.3           & 198            & \multicolumn{1}{l|}{735}            & \multicolumn{1}{l|}{-}           &          -      &       -         &        -        &          -      &  -              \\

\multicolumn{1}{c}{\cellcolor{gray!70}}                                   & \multicolumn{1}{|l|}{+A}            & 75.8 (+0.2)    & 90.1 (+0.7)    & 83.4 (+0.1)    & 201            & \multicolumn{1}{l|}{551}            & \multicolumn{1}{l|}{TraDeS~\cite{wu2021track}}      & 73.7           & 87.1           & 82.3           & 983            & 1,967          \\

\multicolumn{1}{c}{\cellcolor{gray!70}}                                   & \multicolumn{1}{|l|}{+A+TOPIC}      & 77.3 (\textcolor[RGB]{0,100,0}{\textbf{+1.7}})    & 91.5 (\textcolor[RGB]{0,100,0}{\textbf{+2.1}})    & 86.2 (\textcolor[RGB]{0,100,0}{\textbf{+2.9}})    & 212            & \multicolumn{1}{l|}{358}            & \multicolumn{1}{l|}{+TOPIC}      & 74.9 (\textcolor[RGB]{0,100,0}{\textbf{+1.2}})    & 88.4 (\textcolor[RGB]{0,100,0}{\textbf{+1.3}})    & 83.0 (+0.7)    & 989            & 1,489           \\

\multicolumn{1}{c}{\cellcolor{gray!70}}                                   & \multicolumn{1}{|l|}{+A+AARM}       & 76.7 (\textcolor[RGB]{0,100,0}{\textbf{+1.1}})    & 90.6 (\textcolor[RGB]{0,100,0}{\textbf{+1.2}})    & 85.6 (\textcolor[RGB]{0,100,0}{\textbf{+2.3}})    & 243            & \multicolumn{1}{l|}{342}            & \multicolumn{1}{l|}{+AARM}       & 74.3 (+0.6)    & 88.3 (\textcolor[RGB]{0,100,0}{\textbf{+1.2}})    & 82.8 (+0.5)    & 1,021          & 1,467          \\
\multicolumn{1}{c}{\multirow{-5}{*}{\cellcolor{gray!70} \rotatebox{90}{\textit{M-Based}}}}          & \multicolumn{1}{|l|}{+A+TOPIC+AARM} & 78.6 (\textcolor[RGB]{0,100,0}{\textbf{+3.0}})    & 92.4 (\textcolor[RGB]{0,100,0}{\textbf{+3.0}})    & 87.7 (\textcolor[RGB]{0,100,0}{\textbf{+4.4}})    & 289            & \multicolumn{1}{l|}{233}            & \multicolumn{1}{l|}{+TOPIC+AARM} & 75.6 (\textcolor[RGB]{0,100,0}{\textbf{+1.9}})    & 88.7 (\textcolor[RGB]{0,100,0}{\textbf{+1.6}})    & 83.5 (\textcolor[RGB]{0,100,0}{\textbf{+1.2}})    & 1,145          & 1,338          \\ \midrule

\multicolumn{1}{c}{\cellcolor{gray!70}}                                   & \multicolumn{1}{|l|}{Location-based tracker}  & 77.5           & 93.2           & 83.9           & 449             & \multicolumn{1}{l|}{998}           & \multicolumn{1}{l|}{-}           &         -       &     -          &   -             &         -       &    -            \\
\multicolumn{1}{c}{\cellcolor{gray!70}}                                   & \multicolumn{1}{|l|}{+A}            & 82.2           & 93.3           & 91.0           & 72             & \multicolumn{1}{l|}{1,802} & \multicolumn{1}{l|}{-}           &      -          &       -         &            -    &       -         &   -             \\

\multicolumn{1}{c}{\cellcolor{gray!70}}                                   & \multicolumn{1}{|l|}{+M}            & 82.4 (+0.2)    & 93.5 (+0.2)    & 91.0 (+0.0)           & 84             & \multicolumn{1}{l|}{1,726}           & \multicolumn{1}{l|}{FairMOT~\cite{zhang2021fairmot}}     & 53.2           & 69.9           & 71.1           & 3,513          & 8,419          \\

\multicolumn{1}{c}{\cellcolor{gray!70}}                                   & \multicolumn{1}{|l|}{+M+A+TOPIC}      & 83.2 (\textcolor[RGB]{0,100,0}{\textbf{+1.2}})    & 95.5 (\textcolor[RGB]{0,100,0}{\textbf{+2.2}})    & 91.3 (+0.3)    & 194            & \multicolumn{1}{l|}{402}            & \multicolumn{1}{l|}{+TOPIC}      & 54.3 (\textcolor[RGB]{0,100,0}{\textbf{+1.1}})    & 71.1 (\textcolor[RGB]{0,100,0}{\textbf{+1.2}})    & 71.8 (+0.7)    & 3,604           & 7,801           \\

\multicolumn{1}{c}{\cellcolor{gray!70}}                                   & \multicolumn{1}{|l|}{+AARM}         & 82.9 (+0.7)    & 95.3 (\textcolor[RGB]{0,100,0}{\textbf{+2.0}})    & 91.1 (+0.1)    & 196            & \multicolumn{1}{l|}{331}            & \multicolumn{1}{l|}{+AARM}       & 54.1 (+0.9)    & 70.6 (+0.7)    & 71.8 (+0.7)    & 3,667          & 7,756          \\
\multicolumn{1}{c}{\multirow{-5}{*}{\cellcolor{gray!70} \rotatebox{90}{\textit{A-Based}}}}          & \multicolumn{1}{|l|}{+M+TOPIC+AARM} & 84.7 (\textcolor[RGB]{0,100,0}{\textbf{+2.5}})    & 96.6 (\textcolor[RGB]{0,100,0}{\textbf{+3.3}})    & 92.5 (\textcolor[RGB]{0,100,0}{\textbf{+1.5}})    & 205            & \multicolumn{1}{l|}{327}            & \multicolumn{1}{l|}{+TOPIC+AARM} & 54.9 (\textcolor[RGB]{0,100,0}{\textbf{+1.7}})    & 71.9 (\textcolor[RGB]{0,100,0}{\textbf{+2.0}})    & 71.9 (+0.8)    & 3,754          & 7,243          \\ \midrule

\belowrulesepcolor{gray!20} 
\rowcolor{gray!20}
\multicolumn{13}{c}{\textbf{BEE24 Test Set (Ours)}}       
\\
\aboverulesepcolor{gray!20} 
\midrule

\multicolumn{1}{c}{\cellcolor{gray!70}}                                   & \multicolumn{1}{|l|}{ByteTrack~\cite{zhang2022bytetrack}}     & 43.2           & 59.2           & 56.8           & 23,343            & \multicolumn{1}{l|}{44,130}          & \multicolumn{1}{l|}{-}           &           -     &       -         &   -             &           -     &    -            \\

\multicolumn{1}{c}{\cellcolor{gray!70}}                                   & \multicolumn{1}{|l|}{+A}            & 43.4 (+0.2)    & 60.0 (+0.8)    & 56.9 (+0.1)    & 24,431            & \multicolumn{1}{l|}{41,669}           & \multicolumn{1}{l|}{TraDeS~\cite{wu2021track}}      & 30.9           & 42.2           & 34.8           & 56,966            & 26,286            \\

\multicolumn{1}{c}{\cellcolor{gray!70}}                                   & \multicolumn{1}{|l|}{+A+TOPIC}      & 45.2 (\textcolor[RGB]{0,100,0}{\textbf{+2.0}})    & 62.3 (\textcolor[RGB]{0,100,0}{\textbf{+3.1}})    & 59.0 (\textcolor[RGB]{0,100,0}{\textbf{+2.2}})    & 25,624            & \multicolumn{1}{l|}{36,620}            & \multicolumn{1}{l|}{+TOPIC}      & 31.3 (+0.4)    & 42.8 (+0.6)    & 35.7 (+0.9)    & 56,978            & 26,321            \\

\multicolumn{1}{c}{\cellcolor{gray!70}}                                   & \multicolumn{1}{|l|}{+A+AARM}       & 44.5 (\textcolor[RGB]{0,100,0}{\textbf{+1.3}})    & 61.3 (\textcolor[RGB]{0,100,0}{\textbf{+2.1}})    & 57.7 (+0.9)    & 24,903            & \multicolumn{1}{l|}{39,098}            & \multicolumn{1}{l|}{+AARM}       & 31.2 (+0.3)    & 42.6 (+0.4)    & 35.5 (+0.7)    & 56,998            & 26,301            \\
\multicolumn{1}{c}{\multirow{-5}{*}{\cellcolor{gray!70} \rotatebox{90}{\textit{M-Based}}}} & \multicolumn{1}{|l|}{+A+TOPIC+AARM} & 45.3 (\textcolor[RGB]{0,100,0}{\textbf{+2.1}})    & 63.2 (\textcolor[RGB]{0,100,0}{\textbf{+4.0}})    & 59.0 (\textcolor[RGB]{0,100,0}{\textbf{+2.2}})    & 26,252            & \multicolumn{1}{l|}{34,517}            & \multicolumn{1}{l|}{+TOPIC+AARM} & 32.5 (\textcolor[RGB]{0,100,0}{\textbf{+1.6}})    & 43.6 (\textcolor[RGB]{0,100,0}{\textbf{+1.4}})    & 36.9 (\textcolor[RGB]{0,100,0}{\textbf{+2.1}})    & 57,089            & 26,276            \\ \midrule

\multicolumn{1}{c}{\cellcolor{gray!70}}                                   & \multicolumn{1}{|l|}{Location-based tracker}  & 37.7           & 58.5           & 50.4           & 33,224            & \multicolumn{1}{l|}{34,755}           & \multicolumn{1}{l|}{-}           &          -      &    -            & -               &       -         &   -             \\
\multicolumn{1}{c}{\cellcolor{gray!70}}                                   & \multicolumn{1}{|l|}{+A}            & 42.1           & 63.9           & 52.8           & 33,617            & \multicolumn{1}{l|}{25,495} & \multicolumn{1}{l|}{-}           &      -          &       -         &            -    &       -         &   -             \\

\multicolumn{1}{c}{\cellcolor{gray!70}}                                   & \multicolumn{1}{|l|}{+M}            & 42.6 (+0.5)    & 64.2 (+0.3)    & 53.0 (+0.8)    & 33,353            & \multicolumn{1}{l|}{25,291}           & \multicolumn{1}{l|}{FairMOT~\cite{zhang2021fairmot}}     & 42.3           & 40.9           & 54.3           & 75,799            & 18,501          \\

\multicolumn{1}{c}{\cellcolor{gray!70}}                                   & \multicolumn{1}{|l|}{+M+A+TOPIC}      & 44.4 (\textcolor[RGB]{0,100,0}{\textbf{+2.3}})    & 65.9 (\textcolor[RGB]{0,100,0}{\textbf{+2.0}})    & 56.2 (\textcolor[RGB]{0,100,0}{\textbf{+3.4}})    & 32,186            & \multicolumn{1}{l|}{24,018}            & \multicolumn{1}{l|}{+TOPIC}      & 43.5 (\textcolor[RGB]{0,100,0}{\textbf{+1.2}})    & 42.7 (\textcolor[RGB]{0,100,0}{\textbf{+1.8}})    & 55.7 (\textcolor[RGB]{0,100,0}{\textbf{+1.4}})    & 75,812            & 18,470           \\

\multicolumn{1}{c}{\cellcolor{gray!70}}                                   & \multicolumn{1}{|l|}{+AARM}         & 44.4 (\textcolor[RGB]{0,100,0}{\textbf{+2.1}})    & 42.5 (\textcolor[RGB]{0,100,0}{\textbf{+1.6}})    & 56.5 (\textcolor[RGB]{0,100,0}{\textbf{+2.2}})    & 32,956            & \multicolumn{1}{l|}{24,493}            & \multicolumn{1}{l|}{+AARM}       & 43.4 (\textcolor[RGB]{0,100,0}{\textbf{+1.1}})    & 42.2 (\textcolor[RGB]{0,100,0}{\textbf{+1.3}})    & 55.6 (\textcolor[RGB]{0,100,0}{\textbf{+1.3}})    & 75,821            & 18,454          \\
\multicolumn{1}{c}{\multirow{-5}{*}{\cellcolor{gray!70} \rotatebox{90}{\textit{A-Based}}}} & \multicolumn{1}{|l|}{+M+TOPIC+AARM} & 46.6 (\textcolor[RGB]{0,100,0}{\textbf{+4.5}})    & 66.7 (\textcolor[RGB]{0,100,0}{\textbf{+2.8}})    & 59.7 (\textcolor[RGB]{0,100,0}{\textbf{+6.9}})    & 33,171            & \multicolumn{1}{l|}{23,691}            & \multicolumn{1}{l|}{+TOPIC+AARM} & 44.6 (\textcolor[RGB]{0,100,0}{\textbf{+2.3}})    & 43.0 (\textcolor[RGB]{0,100,0}{\textbf{+2.1}})    & 56.8 (\textcolor[RGB]{0,100,0}{\textbf{+2.5}})    & 75,872   & 18,480    \\
\bottomrule
\end{tabular}}
\end{table*}

\subsubsection{Ablation studies of the TOPIC and AARM}
We focus on evaluating the effectiveness of the proposed TOPIC and AARM toward complex and diverse scenes. We compare the existing association paradigms with our proposed parallel association paradigm using four trackers on five datasets, as shown in Table~\ref{tab:ablation}. To thoroughly analyze the contributions of different components, we design a series of ablation studies with various configurations.

For the baseline trackers, we consider both single-feature and serial association paradigms. In the single-feature association paradigm, we include a location-based tracker that uses Euclidean distance between object positions for association without incorporating appearance or advanced motion models. We also use ByteTrack~\cite{zhang2022bytetrack} to represent methods that rely solely on motion features for association.

In the serial association paradigm, we adopt TraDeS~\cite{wu2021track} and FairMOT~\cite{zhang2021fairmot}. TraDeS utilizes motion features as the primary assignment metric and incorporates appearance features sequentially, while FairMOT employs appearance features as the primary assignment metric, assisted by motion features.

Our proposed configurations include various combinations of an appearance model (``A'' in Table~\ref{tab:ablation}), a motion model (``M'' in Table~\ref{tab:ablation}), TOPIC, and AARM.
Specifically, ``A'' denotes incorporating an appearance model (from FastReID~\cite{he2020fastreid}) to extract appearance features to assist motion cues for association. That is, first using the appearance cost matrix to filter some unreliable matches, and then using the cost matrix of the localization-based distance measure or motion model (from ByteTrack or OC-SORT) for association.
``M'' replaces the simple location-based tracker with the OC-SORT~\cite{cao2023observation} motion model for association. 

We also examine combined configurations to assess the contributions of different components in our framework (see Table~\ref{tab:ablation}). The configuration ``A+TOPIC'' augments ByteTrack by adding an appearance model using FastReID to extract appearance features, and employs our proposed parallel association paradigm, TOPIC, for association. 
Building upon configuration ``A'', the ``A+AARM'' setup enhances the appearance model with AARM. In ``M+A+TOPIC'', we discard the location-based metric and instead combine the motion model OC-SORT with the appearance model FastReID for parallel association using TOPIC. 
The configuration ``M+TOPIC+AARM'' extends ``M+A+TOPIC'' by incorporating AARM to enhance the appearance model further. 
Similarly, ``A+TOPIC+AARM'' builds upon ``A+TOPIC'' by adding AARM to the appearance model. Lastly, in ``TOPIC+AARM'', we enhance the appearance models of TraDeS and FairMOT with AARM and utilize TOPIC for association.

\noindent \textbf{Effectiveness of the parallel association paradigm (TOPIC).}
According to Table~\ref{tab:ablation}, we can find that:
(1) the location-based association paradigm is far behind the other paradigms on the HOTA and IDF1 metrics, indicating its poor ability to associate objects and maintain consistency of identities. 
Specifically, the method struggles with tracking consistency in dynamic scenes, where objects move rapidly or change direction. The primary issue is its reliance solely on location-based cues, without considering appearance or motion models. This limitation results in higher association errors and more frequent identity switches;
(2) the serial association paradigm (configurations ``A'' and ``M'') slightly outperforms the single-feature association paradigm (ByteTrack and the Location-based tracker), suggesting that utilizing more features improves tracking performance. This improvement arises because the serial association paradigm combines two features with strengths and weaknesses. These features can complement each other to some extent, helping to reduce tracking errors. Additionally, this result challenges ByteTrack's assertion that a motion-only tracker is superior to a two-feature tracker, as the baselines used in ByteTrack do not employ the same object detector;
(3) compared with other paradigms, our proposed parallel association paradigm (configurations ``A+TOPIC'' and ``TOPIC'') achieves more than a +1.0 improvement in most key metrics (including HOTA, MOTA, and IDF1) across all five datasets. 
As for the serial association paradigm, one feature is used to filter out unreliable matches. However, this can lead to erroneous filtering, especially when a feature, such as appearance, becomes unreliable due to severe occlusion. 
In contrast, our proposed parallel association paradigm adaptively selects the more reliable matching results based on the object's motion level. This heuristic method gives full play to the respective advantages of different features, which reduces the error and improves the overall tracking accuracy.

\begin{figure}
    \centering
    \includegraphics[width=0.4\textwidth]{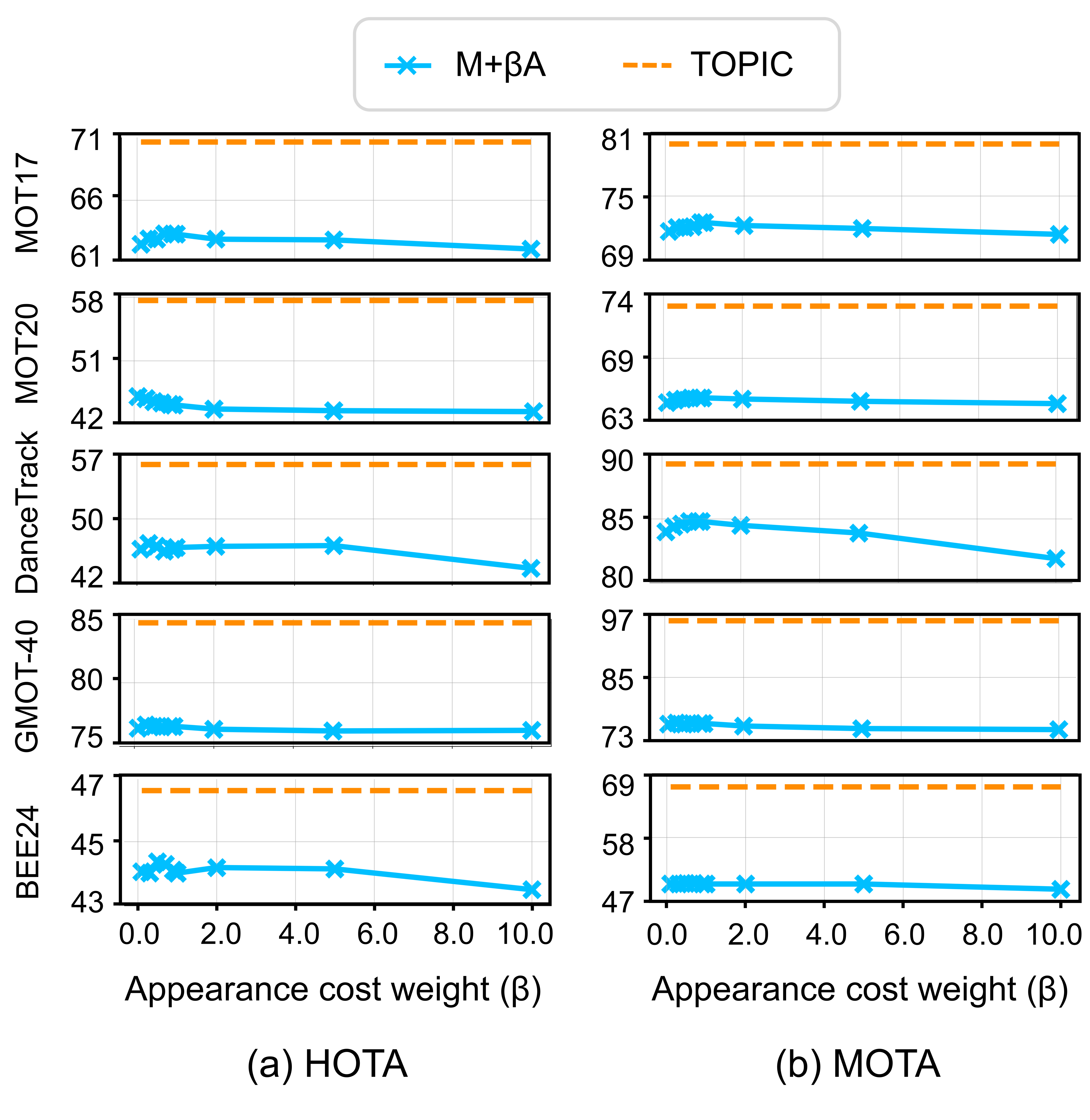}
    \vspace{-0.4cm}
    \hspace{1cm}
    \caption{Comparison of methods for combining appearance and motion features. ``M+$\beta$A'' represents the weighted combination of motion matching cost (M) and appearance matching cost (A) with a weighting factor $\beta$. ``TOPIC'' denotes our two-round parallel matching mechanism. The values of $\beta \in \{0.1, 0.3, 0.5, 0.7, 0.9, 1.0, 2.0, 5.0, 10.0\}$.
    }
    \label{fig:appr_beta}
\end{figure}

To further explore the effectiveness of TOPIC, we employ a straightforward scheme to combine the two features for association by adding the appearance cost matrix and the motion cost matrix.
In order to mitigate scale differences in the costs of different features, we weigh the appearance cost matrix and experiment with a range of parameter values $\{0.1, 0.3, 0.5, 0.7, 0.9, 1.0, 2.0, 5.0, 10.0\}$, as shown in Figure~\ref{fig:appr_beta}.
Experimental results show that such a simple combination leads to worse performance in all metrics on all five datasets.
This is because the essence of the weighting scheme assumes that the cost contribution of appearance and motion features to the total cost is linear.
However, their contribution is always nonlinear.
This causes prone to wrong matching when there is a conflict for the matching considered by the two features, which is common in various scenarios, such as slow motion with occlusion, appearance similarity with slow motion, and fast motion with complete appearance.
In contrast, TOPIC takes into account the applicable scenarios of appearance and motion features, and is able to better utilize the strengths of both features to solve conflicting problems.

\noindent {\textbf{Effectiveness of the AARM.}}
We further validate the effect of AARM on tracking performance, and the results are reported in Table~\ref{tab:ablation}, including configurations ``A+AARM'', ``A+TOPIC+AARM'', ``AARM'', ``M+TOPIC+AARM'', and ``TOPIC+AARM''.
Experimental results show that the AARM consistently improves almost all metrics in different trackers.
This implies that the AARM has a well-generalized ability to enhance the performance of existing tracers.
Moreover, we conclude that assembling TOPIC and AARM at the same time can help different trackers achieve significant performance improvement, such as achieving a +3.3 improvement in HOTA, +4.6 in MOTA, and +3.9 in IDF1 on MOT17 for ByteTrack.
This validates the superiority of our proposed approaches toward complex motions and diverse scenes.

\subsubsection{Visualization results of the TOPIC and AARM} In this subsection, 
we visualize the results of TOPIC and AARM to explore how they improve the tracking performance.

\begin{figure}[!t]
\setlength{\abovecaptionskip}{1pt}
\setlength{\belowcaptionskip}{1pt}
\centering
\includegraphics[width=0.42\textwidth]{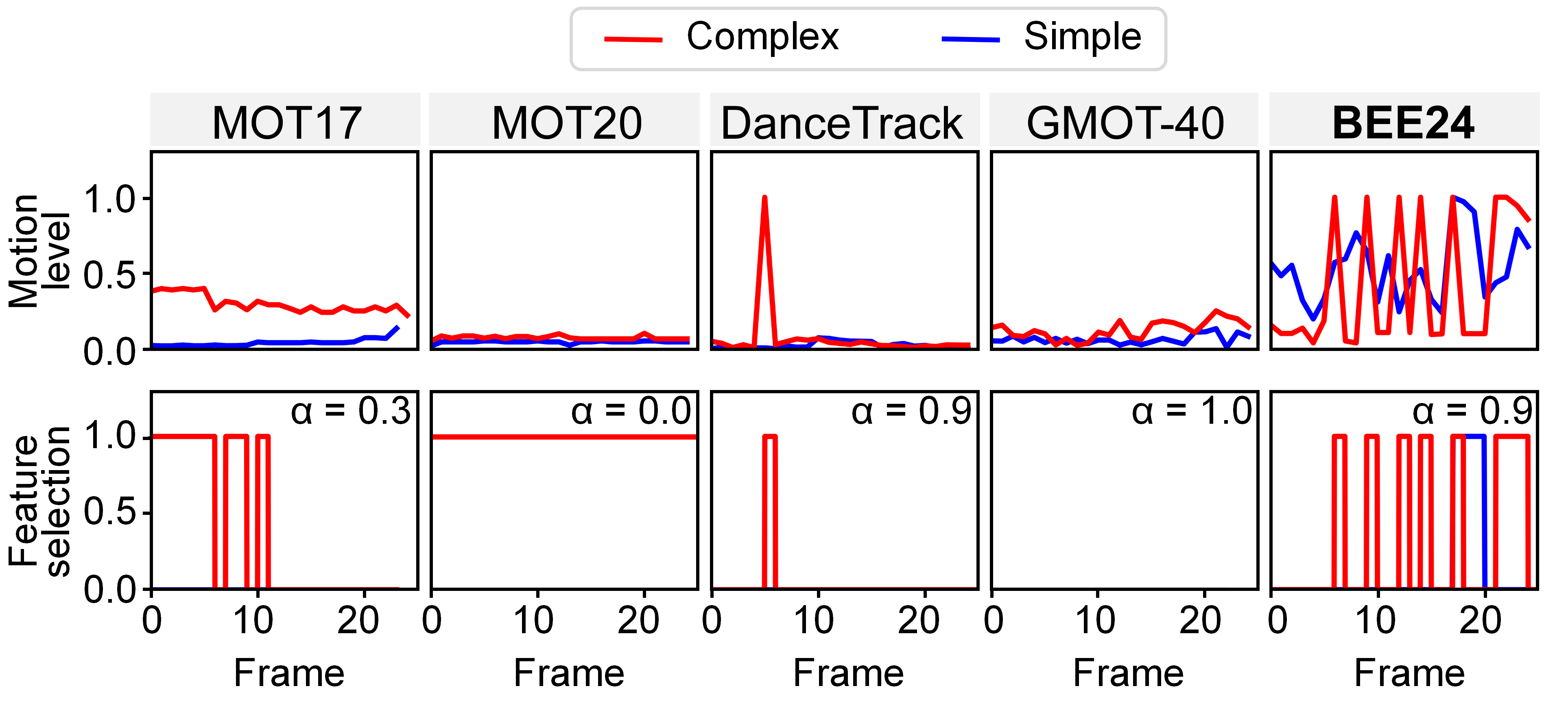}
\hspace{0.3cm}
\vspace{-0.1cm}
\caption{
Visualization of the operation mechanism of the TOPIC on five datasets.
The first row shows the motion level of two objects with different motion patterns within a one-second period;
the second row shows the feature selection of two objects for assignment metrics under the motion level condition in the first row.
$\alpha$ in the upper right corner of the subfigure denotes the motion level thresholds for the dataset.
In the legend, ``Complex'' and ``Simple'' denote the objects with the most complex and simplest motion patterns in the scene, respectively.
}
\label{fig:vis_two-round}
\end{figure}

\noindent {\textbf{Visualization results of the TOPIC.}}
We visualize the matching mechanism of TOPIC to analyze the factors contributing to its superior performance.
As shown in Figure~\ref{fig:vis_two-round}, each dataset compares the motion level and feature selection of simple and complex objects, at around one second.
For feature selection, 0 and 1 denote motion and appearance features, respectively, while $\alpha$ represents the motion level threshold.
We conclude that the TOPIC can adaptively switch appearance or motion features for association according to the current motion level.
This implies that just by adjusting the motion level threshold $\alpha$ appropriately for specific datasets, it is possible to leverage the strengths of both appearance and motion features effectively.

\noindent {\textbf{Visualization results of the AARM.}}
To further explore the impact of AARM on tracking performance, we evaluate the effect of AARM on object appearance representation.
Firstly, we introduce two metrics to analyze the effect of appearance models on object appearance representation.

\noindent \textit{Inter-Class Similarity~\text{(InterCS)}}: 
a metric to measure the average cosine distance of re-ID embeddings among objects in each frame of given videos.
A greater value indicates the smaller the distinction between the representations of different objects' appearance and vice versa.
The value range is $[0, 1]$.
Given a dataset with $V$ videos, \text{InterCS} is formulated as follows:
\begin{align}
\text{InterCS}=
\frac{1}{\sum_{v=1}^V T_{v}} \sum_{v=1}^V \sum_{t=1}^{T_{v}} \frac{1}{N_{v,t}^2} \sum_i^{N_{v,t}} \sum_{j \neq i}^{N_{v,t}}\cos (E_{v,t}^{i},E_{v,t}^{j}).
\end{align}
where $T_{v}$ represents the number of frames of the $v$-th video, $N_{v,t}$ represents the number of objects in the $t$-th frame of video $v$,  $E_{v,t}^{i}$ and $E_{v,t}^{j}$ denote appearance embedding of the object $i$ and $j$ in the $t$-th frame of video $v$, respectively, and $\cos$ is the function to calculate cosine distance between two embeddings.

\noindent \textit{Intra-Class Similarity~\text{(IntraCS)}}: 
A measure of the average cosine distance of re-ID embeddings of a single object across frames of given videos.
A greater value indicates a greater similarity in the representation of a single object's appearance across frames and vice versa.
The value range is $[0, 1]$.
Given a dataset with $V$ videos, IntraCS is formulated as follows:

\begin{align}
\text{IntraCS}=\frac{1}{\sum_{v=1}^V n_{v}} \sum_{v=1}^V \sum_{i=1}^{n_{v}} \frac{1}{L_{v,i}^{2}} \sum_k^{L_{v,i}} \sum_{q \neq k}^{L_{v,i}}\cos (E_{v,k}^{i},E_{v,q}^{i}).
\end{align}
where $n_{v}$ denotes the number of objects in the $v$-th video, $L_{v,i}$ represents the duration of frames of object $i$ of video $v$,  $E_{v,k}^{i}$ and $E_{v,q}^{i}$ denote appearance embedding of object $i$ in the frame $k$ and $q$ of video $v$, respectively.
Note that objects include deleted ones during the tracking process and the remaining ones at the end of each video.

\begin{figure}[!t]
\centering
\includegraphics[width=0.4\textwidth]{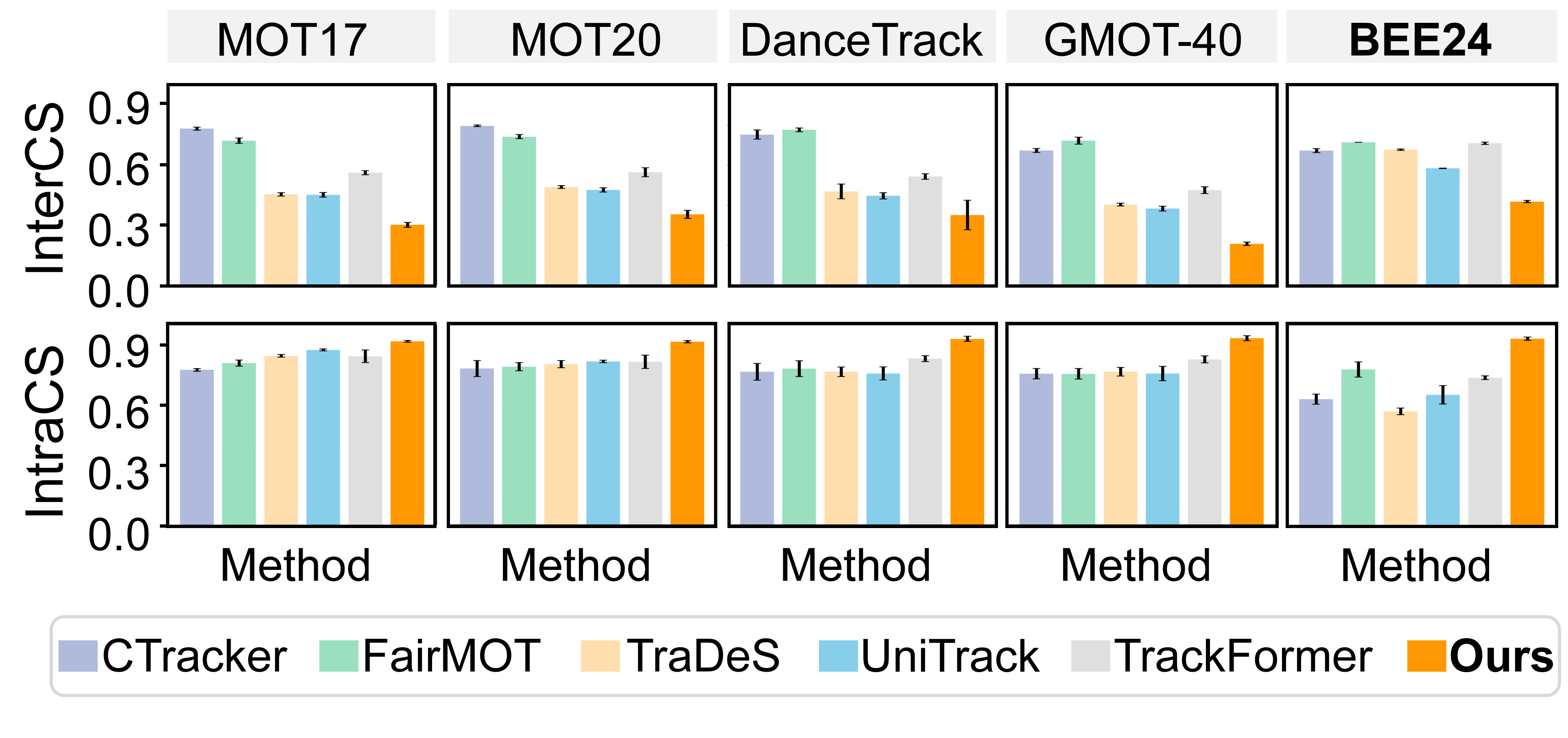}
\vspace{-0.5cm}
\caption{
Comparison of appearance representation quality of different trackers on five datasets.
The first row is InterCS, a smaller value indicates the greater distinction between the representations of different objects’ appearance;
the second row is IntraCS, a greater value indicates a greater similarity in representing a single object’s appearance across frames.
}
\label{fig:vs_emb}
\end{figure}

We compare the appearance representation capability of our proposed TOPICTrack with five state-of-the-art trackers which use private detectors and re-ID modules, including CTracker~\cite{peng2020chained}, FairMOT~\cite{zhang2021fairmot}, TraDeS~\cite{wu2021track}, UniTrack~\cite{wang2021different}, and TrackFormer~\cite{meinhardt2022trackformer}.
The comparison results are reported in Figure~\ref{fig:vs_emb}, which show that our method achieves the smallest inter-class similarity and highest intra-class similarity on all five datasets.
The experimental results demonstrate the ability of the AARM to enhance the ability to distinguish among objects and the consistency of the same object across frames.

\begin{figure}[!h]
\setlength{\abovecaptionskip}{1pt}
\setlength{\belowcaptionskip}{1pt}
    \centering
    \includegraphics[width=0.8\linewidth]{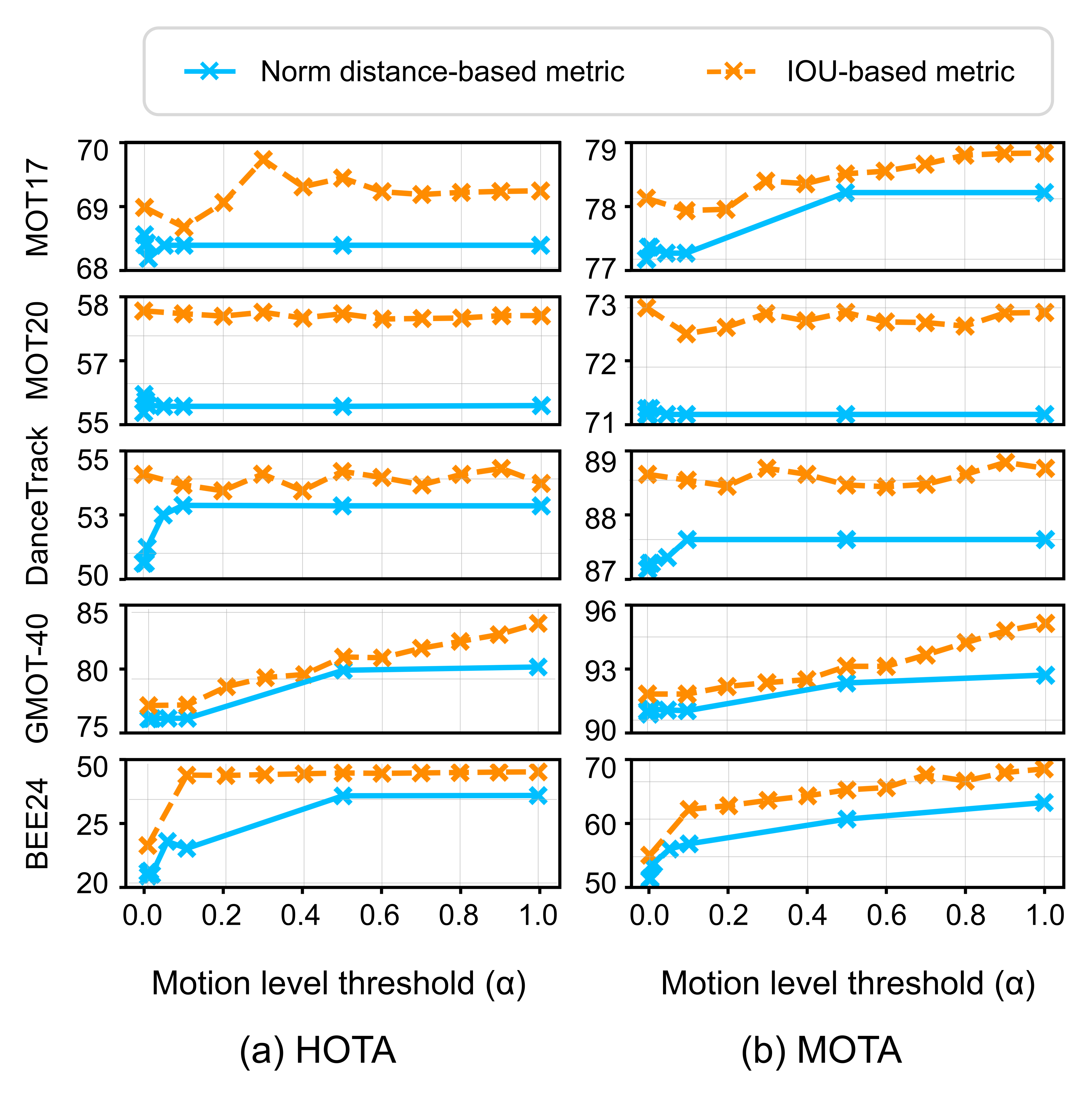}
    \vspace{-0.35cm}
    \hspace{0.8cm}
    \caption{
    Sensitivity analysis of the motion level threshold $\alpha$, which is useful when there are conflicting matches.
    (a) and (b) represent the effect of $\alpha$ with respect to HOTA and MOTA on the five datasets, respectively.
    ``Norm distance-based metric'' with parameter range $\alpha \in \{0, 0.001, 0.005, 0.01, 0.05, 0.1, 0.5, 1\}$.
    ``IOU-based metric'' with parameter range $\alpha \in \{0.1, 0.2, 0.3, 0.4, 0.5, 0.6, 0.7, 0.8, 0.9, 1\}$.
    }
    \label{fig:sensitivity analysis}
\end{figure}

\subsubsection{Sensitivity analysis of the threshold of motion level}
The motion level threshold $\alpha$ is handy when conflict matches occur.
For the purpose of exploring the impact of different values of the motion level threshold $\alpha$ on the tracking performance, we conduct a sensitivity analysis on this parameter.
On all five datasets, we perform 10 separate group experiments via grid search, where $\alpha \in [0,1]$ with an interval of 0.1.

As shown in Figure~\ref{fig:sensitivity analysis}, we observe that the threshold of the motion level affects the tracking performance, and the degree of the effect is correlated with the conflict rate.
For instance, GMOT-40 has a conflict rate of 15.4\%, which varies around 6 in both HOTA and MOTA.
We also note that DanceTrack's conflict rate, although also high (13.3\%), the motion level threshold $\alpha$ has a smaller impact on tracking performance.
We analyze the reason that many motion and appearance matches are wrong, the association results will be wrong regardless of what kind of feature is selected.

Besides, compared to using IOU as a metric, we introduce the normalized center distance between bounding boxes as a metric. Specifically, we use the length and width of the image to normalize the distance between objects.
Since most of the normalized distances are smaller than IOU-based distances, we set a finer interval for the motion level threshold $\alpha$, i.e., $\alpha \in \{0, 0.001, 0.005, 0.01, 0.05, 0.1, 0.5, 1\}$. 
As shown in Figure~\ref{fig:sensitivity analysis}, both the HOTA and MOTA using normalized center distance as the motion level metric are lower than the best accuracy of the IOU-based metrics on all five datasets.
We argue that this is because normalized center distance ignores the scale differences of different objects, thereby unsuitable as a unified motion level metric. In contrast, IOU normalizes the motion level for each object individually by calculating the overlapping area of bounding boxes of two successive frames, which is more suitable as a unified motion level metric.

\subsubsection{Computational analysis of Re-ID module}
To assess the impact of the Re-ID module on the computational speed of our tracker, we conduct experiments using the TOPICTrack on five MOT datasets: MOT17, MOT20, DanceTrack, GMOT-40, and BEE24. 
We compare the computation time (in milliseconds) and accuracy (measured by the HOTA metric) of the algorithm in two configurations: with the Re-ID module included (i.e., TOPICTrack) and with the Re-ID module removed (i.e., using only the motion model).

\begin{figure}[!t]
\setlength{\abovecaptionskip}{1pt}
\setlength{\belowcaptionskip}{1pt}
\centering
\includegraphics[width=0.35\textwidth]{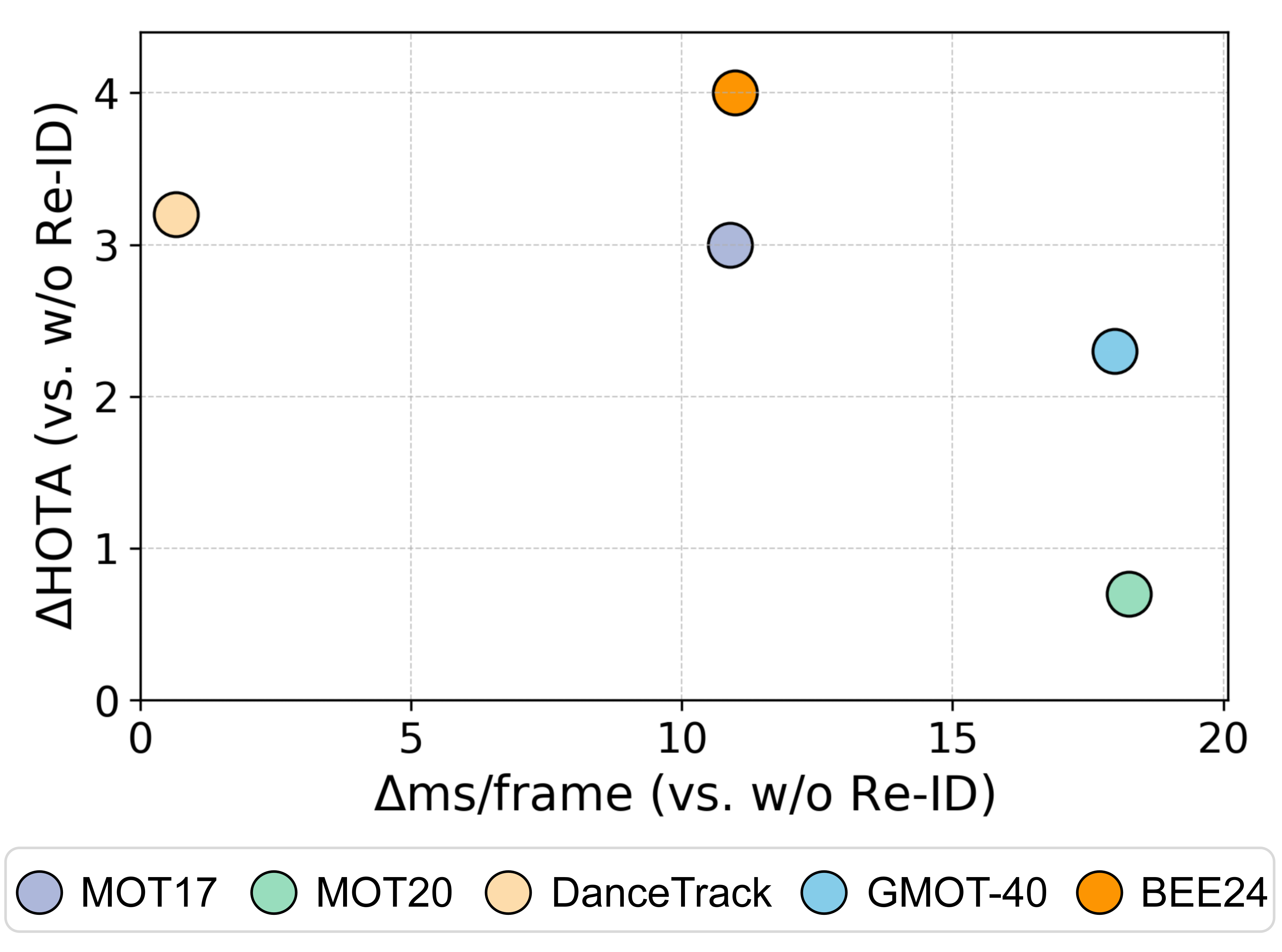}
\hspace{0.3cm}
\vspace{-0.2cm}
\caption{
Impact of the Re-ID module on computational time and accuracy across five MOT datasets. 
The x-axis represents the increase in computation time per frame ($\Delta$ms) after adding the Re-ID module, and the y-axis shows the corresponding accuracy improvement ($\Delta$HOTA).
}
\label{fig:FPS_HOTA}
\end{figure}

As shown in Figure~\ref{fig:FPS_HOTA}, we observe that incorporating the Re-ID module led to accuracy improvements across all datasets, with $\Delta$HOTA ranging from 0.7\% to 4.0\%. Moreover, the accuracy gains are more significant in scenes with more complex motion patterns (see Figure~\ref{fig:MMSAO and MMSO}). The increase in computation time is relatively small, with $\Delta$ms ranging from 1~ms to 18~ms, and it is higher in scenes with more objects (see Table~\ref{tab:comparsion_dataset}). Although the Re-ID module introduces some computational overhead, this increase is acceptable in practical applications. This analysis confirms that our algorithm is both effective and practical.

\subsection{Benchmark Evaluation}

\noindent \textbf{MOT17 and MOT20.} We compare the proposed TOPICTrack tracker with the state-of-the-art trackers on the MOT17 and MOT20 test sets.
Since we do not use the public detection results, the ``private detector'' protocol is adopted. Note that all of the results are directly obtained from the official MOT challenge evaluation server.
As shown in Table~\ref{tab:benchmark_mot17} and Table~\ref{tab:benchmark_mot20}, our TOPICTrack tracker achieves the best results on most metrics, e.g., HOTA, IDF1, IDs, AssA, AssR, etc. 

\noindent \textbf{DanceTrack}. As shown in Table~\ref{tab:benchmark_dance}, we compare TOPICTrack with the state-of-the-art trackers on DanceTrack test sets. Table~\ref{tab:benchmark_dance} shows that the TOPICTrack tracker outperforms other trackers by a large margin on most metrics.
Specifically, the TOPICTrack is led by 0.9 HOTA, 1.6 AssA. This suggests that our tracker set a new state-of-the-art.

\begin{table*}[!ht]
\caption{Results on MOT17 test set with the private detections.
Methods in the bottom \textcolor{gray}{gray} block use the same detections.
$\uparrow$ follows the metric indicating the larger value, the better performance, and vice versa.
\textbf{Bolding} and \underline{underline} denote the best and second-best results in each column, respectively.
}
\label{tab:benchmark_mot17}
\centering
\setlength{\tabcolsep}{5.5pt}
\scriptsize
\scalebox{0.85}{
\begin{tabular}{llllllllll}
\toprule
\multicolumn{1}{l|}{Tracker}                           & HOTA$\uparrow$  & MOTA$\uparrow$ & IDF1$\uparrow$ & AssA$\uparrow$ & AssR$\uparrow$ & FP($10^{4}$)$\downarrow$  & FN($10^{4}$)$\downarrow$    & IDs$\downarrow$  & Frag$\downarrow$ \\ \midrule
\multicolumn{1}{l|}{FairMOT~\cite{zhang2021fairmot}}                           & 59.3 & 73.7 & 72.3 & 58.0 & 63.6 & 2.75 & 11.70 & 3,303  & 8,073 \\
\multicolumn{1}{l|}{TransCt~\cite{xu2021transcenter}}                           & 54.5 & 73.2 & 62.2 & 49.7 & 54.2 & 2.31 & 12.40 & 4,614  & 9,519 \\
\multicolumn{1}{l|}{TransTrk~\cite{sun2020transtrack}}                          & 54.1 & 75.2 & 63.5 & 47.9 & 57.1 & 5.02 & 8.64  & 3,603  & 4,872 \\
\multicolumn{1}{l|}{GRTU~\cite{wang2021general}}                              & 62.0 & 74.9 & 75.0 & 62.1 & 65.8 & 3.20 & 10.80 & \underline{1,812}  & \textbf{1,824} \\
\multicolumn{1}{l|}{QDTrack~\cite{pang2021quasi}}                           & 53.9 & 68.7 & 66.3 & 52.7 & 57.2 & 2.66 & 14.70 & 3,378  & 8,091 \\
\multicolumn{1}{l|}{MOTR~\cite{zeng2022motr}}                              & 57.2 & 71.9 & 68.4 & 55.8 & 59.2 & 2.11 & 13.60 & 2,115  & 3,897 \\
\multicolumn{1}{l|}{PermaTr~\cite{tokmakov2021learning}}                           & 55.5 & 73.8 & 68.9 & 53.1 & 59.8 & 2.90 & 11.50 & 3,699  & 6,132 \\
\multicolumn{1}{l|}{TransMOT~\cite{chu2023transmot}}                          & 61.7 & 76.7 & 75.1 & 59.9 & 66.5 & 3.62 & 9.32  & 2,346  & 7,719 \\
\multicolumn{1}{l|}{GTR~\cite{zhou2022global}}                               & 59.1 & 75.3 & 71.5 & 61.6 & - & 2.68 & 11.00 & 2,859     & -    \\
\multicolumn{1}{l|}{DST-Tracker~\cite{cao2022track}}                       & 60.1 & 75.2 & 72.3 & 62.1 & - & 2.42 & 11.00 & 2,729     & -    \\
\multicolumn{1}{l|}{MeMOT~\cite{cai2022memot}}                             & 56.9 & 72.5 & 69.0 & 55.2 & - & 2.72 & 11.50 & 2,724     & -    \\
\multicolumn{1}{l|}{UniCorn~\cite{yan2022towards}}                           & 61.7 & 77.2 & 75.5 & -    & - & 5.01 & \textbf{7.33}  & 5,379     & -    \\
\multicolumn{1}{l|}{STONet-TIA~\cite{li2023inference}}                           & 59.5 & 76.0 & 71.9 & -    & - & 2.75 & 10.51  & 2,784     & -    \\
\multicolumn{1}{l|}{Decode-MOT~\cite{lee2023decode}}                           & 59.6 & 73.2 & 72.0 & -    & - & 2.65 & 12.13  & 3,363     & -    \\
\multicolumn{1}{l|}{QDTrackV2~\cite{fischer2023qdtrack}}                           & 63.5 & 78.7 & 77.5 & 62.6    & 65.7 & - & -  & -     & -    \\
\multicolumn{1}{l|}{FastTrack~\cite{liu2023fasttrack}}                           & 63.4 & \underline{80.5} & 77.6 & 62.3    & 67.8 & 2.47 & 8.36  & 2,013     & 2,337    \\
\multicolumn{1}{l|}{BPMTrack~\cite{gao2024bpmtrack}}                           & \underline{63.6} & \textbf{81.3} & \underline{78.1} & -    & - & 2.58 & \underline{7.79}  & 2,021     & -    \\ \midrule
\belowrulesepcolor{gray!10} 
\rowcolor{gray!10}
\multicolumn{1}{l|}{ByteTrack~\cite{zhang2022bytetrack}} & 63.1 & 80.3 & 77.3 & 62.0 & \underline{68.2} & 2.55 & 8.37  & 2,196  & 2,277 \\
\rowcolor{gray!10}
\multicolumn{1}{l|}{OC-SORT~\cite{cao2023observation}}   & 63.2 & 78.0 & 77.5 & \underline{63.2} & 67.5 & \textbf{1.51} & 10.80 & 1,950  & \underline{2,040} \\
\rowcolor{gray!10}
\multicolumn{1}{l|}{\textbf{TOPICTrack (Ours)}}      & \textbf{63.9} & 78.8 & \textbf{78.7} & \textbf{64.3} & \textbf{69.9} & \underline{1.70} & 10.11 & \textbf{1,515}  &  2,613    \\
\aboverulesepcolor{gray!10}
\bottomrule
\end{tabular}}
\end{table*}

\begin{table*}[]
\caption{Results on MOT20 test set with the private detections.
Methods in the bottom \textcolor{gray}{gray} block use the same detections.
$\uparrow$ follows the metric indicating the larger value, the better performance, and vice versa.
\textbf{Bolding} and \underline{underline} denote the best and second-best results in each column, respectively.
}
\label{tab:benchmark_mot20}
\centering
\setlength{\tabcolsep}{5.5pt}
\scriptsize
\scalebox{0.85}{
\begin{tabular}{llllllllll}
\toprule
\multicolumn{1}{l|}{Tracker}                           & HOTA$\uparrow$  & MOTA$\uparrow$ & IDF1$\uparrow$ & AssA$\uparrow$ & AssR$\uparrow$ & FP($10^{4}$)$\downarrow$  & FN($10^{4}$)$\downarrow$    & IDs$\downarrow$  & Frag$\downarrow$ \\ \midrule
\multicolumn{1}{l|}{FairMOT~\cite{zhang2021fairmot}}                           & 54.6 & 61.8 & 67.3 & 54.7 & 60.7 & 10.30 & 8.89  & 5,243  & 7,874 \\
\multicolumn{1}{l|}{TransCt~\cite{xu2021transcenter}}                           & 43.5 & 58.5 & 49.6 & 37.0 & 45.1 & 6.42  & 14.60 & 4,695  & 9,581 \\
\multicolumn{1}{l|}{Semi-TCL~\cite{li2021semitcl}}                          & 55.3 & 65.2 & 70.1 & 56.3 & 60.9 & 6.12  & 11.50 & 4,139  & 8,508 \\
\multicolumn{1}{l|}{CSTrack~\cite{liang2022rethinking}}                           & 54.0 & 66.6 & 68.6 & 54.0 & 57.6 & 2.54  & 14.40 & 3,196  & 7,632 \\
\multicolumn{1}{l|}{GSDT~\cite{wang2021joint}}                              & 53.6 & 67.1 & 67.5 & 52.7 & 58.5 & 3.19  & 13.50 & 3,131  & 9,875 \\
\multicolumn{1}{l|}{TransMOT~\cite{chu2023transmot}}                          & 61.9 & 77.5 & 75.2 & 60.1 & 66.3 & 3.42  & \textbf{8.08}  & 1,615  & 2,421 \\
\multicolumn{1}{l|}{MeMOT~\cite{cai2022memot}}                             & 54.1 & 63.7 & 66.1 & 55.0 & - & 4.79  & 13.80 & 1,938     & -    \\

\multicolumn{1}{l|}{STONet-TIA~\cite{li2023inference}}                           & 55.2 & 69.4 & 65.8 & -    & - & 6.24 & 11.08  & 3,975     & -    \\
\multicolumn{1}{l|}{Decode-MOT~\cite{lee2023decode}}                           & 54.5 & 67.2 & 69.0 & -    & - & 3.52 & 13.15  & 2,805     & -    \\
\multicolumn{1}{l|}{QDTrackV2~\cite{fischer2023qdtrack}}                           & 60.0 & 74.7 & 73.8 & 58.9    & 61.4 & - & -  & -     & -    \\
\multicolumn{1}{l|}{FastTrack~\cite{liu2023fasttrack}}                           & 61.8 & 77.3 & 74.7 & 60.2    & 66.9 & 2.53 & 9.06  & 1,434     & \underline{1,337}    \\
\multicolumn{1}{l|}{BPMTrack~\cite{gao2024bpmtrack}}                           &\underline{62.3} & \textbf{78.3} & \underline{76.7} & -    & - & 2.86 & \underline{8.25}  & 1,314     & -    \\

\midrule
\belowrulesepcolor{gray!10} 
\rowcolor{gray!10}
\multicolumn{1}{l|}{ByteTrack~\cite{zhang2022bytetrack}} & 61.3 & \underline{77.8} & 75.2 & 59.6 & 66.2 & 2.62  & 8.76  & 1,223  & 1,460 \\
\rowcolor{gray!10} 
\multicolumn{1}{l|}{OC-SORT~\cite{cao2023observation}}   & 62.1 & 75.5 & 75.9 & \underline{62.0} & \underline{67.5} & \underline{1.80}  & 10.80 & \underline{913}   & \textbf{1,198} \\
\rowcolor{gray!10}
\multicolumn{1}{l|}{\textbf{TOPICTrack (Ours)}}      & \textbf{62.6} & 72.4 & \textbf{77.6} & \textbf{65.4} & \textbf{70.3} & \textbf{1.10}  & 13.11 & \textbf{869}   & 1,574    \\ 
\aboverulesepcolor{gray!10}
\bottomrule
\end{tabular}}
\end{table*}

\begin{table}[!h]
\caption{Results on DanceTrack test set with the private detections.
Methods in the bottom \textcolor{gray}{gray} block use the same detections.
$\uparrow$ follows the metric indicating the larger value, the better performance, and vice versa.
\textbf{Bolding} and \underline{underline} denote the best and second-best results in each column, respectively.
}
\label{tab:benchmark_dance}
\centering
\setlength{\tabcolsep}{3pt}
\scriptsize
\scalebox{0.85}{
\begin{tabular}{llllll}
\toprule

\multicolumn{1}{l|}{Tracker}     & HOTA$\uparrow$ & DetA$\uparrow$ & AssA$\uparrow$ & MOTA$\uparrow$ & IDF1$\uparrow$ \\ \midrule
\multicolumn{1}{l|}{CenterTrack~\cite{zhou2020tracking}} & 41.8 & 78.1 & 22.6 & 86.8 & 35.7 \\
\multicolumn{1}{l|}{FairMOT~\cite{zhang2021fairmot}}     & 39.7 & 66.7 & 23.8 & 82.2 & 40.8 \\
\multicolumn{1}{l|}{QDTrack~\cite{pang2021quasi}}     & 45.7 & 72.1 & 29.2 & 83.0 & 44.8 \\
\multicolumn{1}{l|}{TransTrk~\cite{sun2020transtrack}}    & 45.5 & 75.9 & 27.5 & 88.4 & 45.2 \\
\multicolumn{1}{l|}{TraDeS~\cite{wu2021track}   }    & 43.3 & 74.5 & 25.4 & 86.2 & 41.2 \\
\multicolumn{1}{l|}{MOTR~\cite{zeng2022motr}}        & 54.2 & 73.5 & 40.2 & 79.7 & 51.5 \\
\multicolumn{1}{l|}{GTR~\cite{zhou2022global}}         & 48.0 & 72.5 & 31.9 & 84.7 & 50.3 \\
\multicolumn{1}{l|}{DST-Tracker~\cite{cao2022track}} & 51.9 & 72.3 & 34.6 & 84.9 & 51.0 \\

\multicolumn{1}{l|}{QDTrackV2~\cite{fischer2023qdtrack}} & 54.2 & 80.1 & 36.8 & 87.7 & 50.4 \\
\multicolumn{1}{l|}{FastTrack~\cite{liu2023fasttrack}} & \underline{57.4} & \textbf{81.1} & \underline{40.7} & \textbf{92.8} & \underline{58.2} \\

\midrule
\belowrulesepcolor{gray!10} 
\rowcolor{gray!10}
\multicolumn{1}{l|}{SORT~\cite{bewley2016simple}}        & 47.9 & 72.0 & 31.2 & 91.8 & 50.8 \\
\rowcolor{gray!10}
\multicolumn{1}{l|}{DeepSORT~\cite{wojke2017simple}}    & 45.6 & 71.0 & 29.7 & 87.8 & 47.9 \\
\rowcolor{gray!10}
\multicolumn{1}{l|}{ByteTrack~\cite{zhang2022bytetrack}}   & 47.3 & 71.6 & 31.4 & 89.5 & 52.5 \\
\rowcolor{gray!10}
\multicolumn{1}{l|}{OC-SORT~\cite{cao2023observation}}     & 55.1 & 80.4 & 40.4 & \underline{92.2} & 54.9 \\
\rowcolor{gray!10}
\multicolumn{1}{l|}{\textbf{TOPICTrack (Ours)}}        & \textbf{58.3} & \underline{80.7} & \textbf{42.3} & 90.9 & \textbf{58.4} \\
\aboverulesepcolor{gray!10}
\bottomrule
\end{tabular}}
\end{table}

\noindent \textbf{GMOT-40}. We compare our approach to the state-of-the-art trackers on GMOT-40 test sets, as seen in Table~\ref{tab:benchmark_gmot}. 
Since there is no official division of training, validation, and test sets, and no leaderboards are provided, we divide the training and test sets according to the scheme mentioned in Section~\ref{sec:Experiment Setup}.
For the other trackers, we train and test them according to their respective open-source code and parameter settings.
The experimental results are shown in Table~\ref{tab:benchmark_gmot}, and our method achieves remarkably leading results on the majority of metrics, especially on HOTA by 2.5, on MOTA by 3.3, on IDF1 by 1.5 and on AssR by 4.6.

\noindent \textbf{BEE24}. Similar to the comparison on GMOT-40, we re-train all trackers on BEE24. In Table~\ref{tab:benchmark_bee24}, the experimental results demonstrate that the proposed TOPICTrack tracker achieves notable improvements in most metrics, e.g., improving HOTA by 2.3, MOTA by 5.1, and IDF1 by 2.9.
We also note that across all five datasets, all algorithms have the lowest accuracy on BEE24, and the AssA metric, which measures detection performance, has a maximum value of only about 40 across all algorithms. This is due to the difficulty of detecting small objects, suggesting that BEE24 increases the challenge of detection and tracking, and poses a more difficult task for existing algorithms.

\begin{table*}[]
\caption{Results on GMOT-40 test set with the private detections.
Methods in the bottom \textcolor{gray}{gray} block use the same detections.
$\uparrow$ follows the metric indicating the larger value, the better performance, and vice versa.
\textbf{Bolding} and \underline{underline} denote the best and second-best results in each column, respectively.
}
\label{tab:benchmark_gmot}
\centering
\setlength{\tabcolsep}{7pt}
\scriptsize
\scalebox{0.85}{
\begin{tabular}{l|lllllllll}
\toprule
\multicolumn{1}{l|}{Tracker} & HOTA$\uparrow$  & MOTA$\uparrow$ & IDF1$\uparrow$ & AssA$\uparrow$ & AssR$\uparrow$ & FP$\downarrow$  & FN$\downarrow$    & IDs$\downarrow$  & Frag$\downarrow$ \\ \midrule
Ctracker~\cite{peng2020chained}                     & 47.3 & 59.9 & 69.4 & 42.1   & 43.5 & 2,632 & 6,743 & 386   & 327  \\
FairMOT~\cite{zhang2021fairmot}                      & 53.2 & 69.9 & 71.1 & 56.3   & 57.6 & 3,513 & 8,419 & 1,973  & 265  \\
TraDeS~\cite{wu2021track}                       & 73.7 & 87.1 & 82.3 & 74.9   & 76.4 & 983  & 1,967 & 1,346   & 189  \\

TrackFormer~\cite{meinhardt2022trackformer}                  & 54.2 & 70.2 & 65.3 & 51.4   & 53.8 & 3,572 & 5,643 & 1,396   & 243  \\ \midrule
\belowrulesepcolor{gray!10} 
\rowcolor{gray!10}
UniTrack~\cite{wang2021different}                     & 56.3 & 72.1 & 70.3 & 54.1   & 55.2 & 3,467 & 5,673 & 395   & 236  \\
\rowcolor{gray!10}
ByteTrack~\cite{zhang2022bytetrack}                    & 75.6 & 89.4 & 83.3 & 77.1   & \underline{80.3} & \underline{198}  & \underline{735}  & \textbf{187}   & \underline{102}  \\
\rowcolor{gray!10}
OC-SORT~\cite{cao2023observation}                      & \underline{82.2} & \underline{93.3} & \underline{91.0} & \underline{81.2}   & 80.2 & \textbf{72}  & 1,802 & \underline{220}   & 152  \\
\rowcolor{gray!10}
\textbf{TOPICTrack (Ours)}                          & \textbf{84.7} & \textbf{96.6} & \textbf{92.5} & \textbf{82.7} & \textbf{84.9} & 205  & \textbf{327}  & 335   & \textbf{92}   \\
\aboverulesepcolor{gray!10}
\bottomrule
\end{tabular}}
\end{table*}

\begin{table*}[!h]
\caption{Results on BEE24 test set with the private detections.
Methods in the bottom \textcolor{gray}{gray} block use the same detections.
$\uparrow$ follows the metric indicating the larger value, the better performance, and vice versa.
\textbf{Bolding} and \underline{underline} denote the best and second-best results in each column, respectively.
}
\label{tab:benchmark_bee24}
\centering
\setlength{\tabcolsep}{7pt}
\scriptsize
\scalebox{0.85}{
\begin{tabular}{l|lllllllll}
\toprule
Tracker     & HOTA$\uparrow$ & MOTA$\uparrow$ & IDF1$\uparrow$           & AssA$\uparrow$ & AssR$\uparrow$ & FP$\downarrow$ & FN$\downarrow$ & IDs$\downarrow$ & Frag$\downarrow$ \\ \midrule
Ctracker~\cite{peng2020chained}    & 33.4         & 42.8           & 39.4                     & 24.5          & 34.0         & 65,532         & 23,983         & 2,987           & 2,770            \\
FairMOT~\cite{zhang2021fairmot}     & 42.3         & 40.9           & 54.3                     & \textbf{42.5}         & \underline{57.6}         & 75,799         & \underline{18,501}         & 3,968           & 3,790            \\
TraDeS~\cite{wu2021track}      & 30.9         & 42.2           & 34.8                     & 20.2         & 26.6         & 56,966         & 26,286         & 5,660           & 4,716            \\
TrackFormer~\cite{meinhardt2022trackformer} & \underline{44.3}         & 41.5           & 53.9                     & \underline{42.3}         & 55.5         & 86,777         & \textbf{9,989}          & 3,405           & \textbf{2,271}            \\ \midrule
\belowrulesepcolor{gray!10} 
\rowcolor{gray!10}
UniTrack~\cite{wang2021different}    & 41.6         & 54.6           & 53.0 & 34.8         & 56.2         & 51,369         & 21,953         & 1,972           & \underline{2,395}            \\
\rowcolor{gray!10}
ByteTrack~\cite{zhang2022bytetrack}   & 43.2         & 59.2           & \underline{56.8}                     & 38.3         & 52.7         & \underline{23,343}         & 44,130         & \textbf{1,303}           & 6,663            \\
\rowcolor{gray!10}
OC-SORT~\cite{cao2023observation}     & 42.7         & \underline{61.6}           & 55.3                     & 36.8         & 50.8         & \textbf{20,493}         & 43,172         & 1,435           & 4,996            \\
\rowcolor{gray!10}
\textbf{TOPICTrack (Ours)}  & \textbf{46.6}          & \textbf{66.7}           & \textbf{59.7}                     & 40.3         & \textbf{59.1}         & 29,171         & 25,691         & \underline{1,401}           & 2,490            \\ \aboverulesepcolor{gray!10}
\bottomrule
\end{tabular}}
\end{table*}

\subsection{Case Study}

\begin{figure}[!h]
\setlength{\abovecaptionskip}{1pt}
\setlength{\belowcaptionskip}{1pt}
    \centering
    \includegraphics[width=0.9\linewidth]{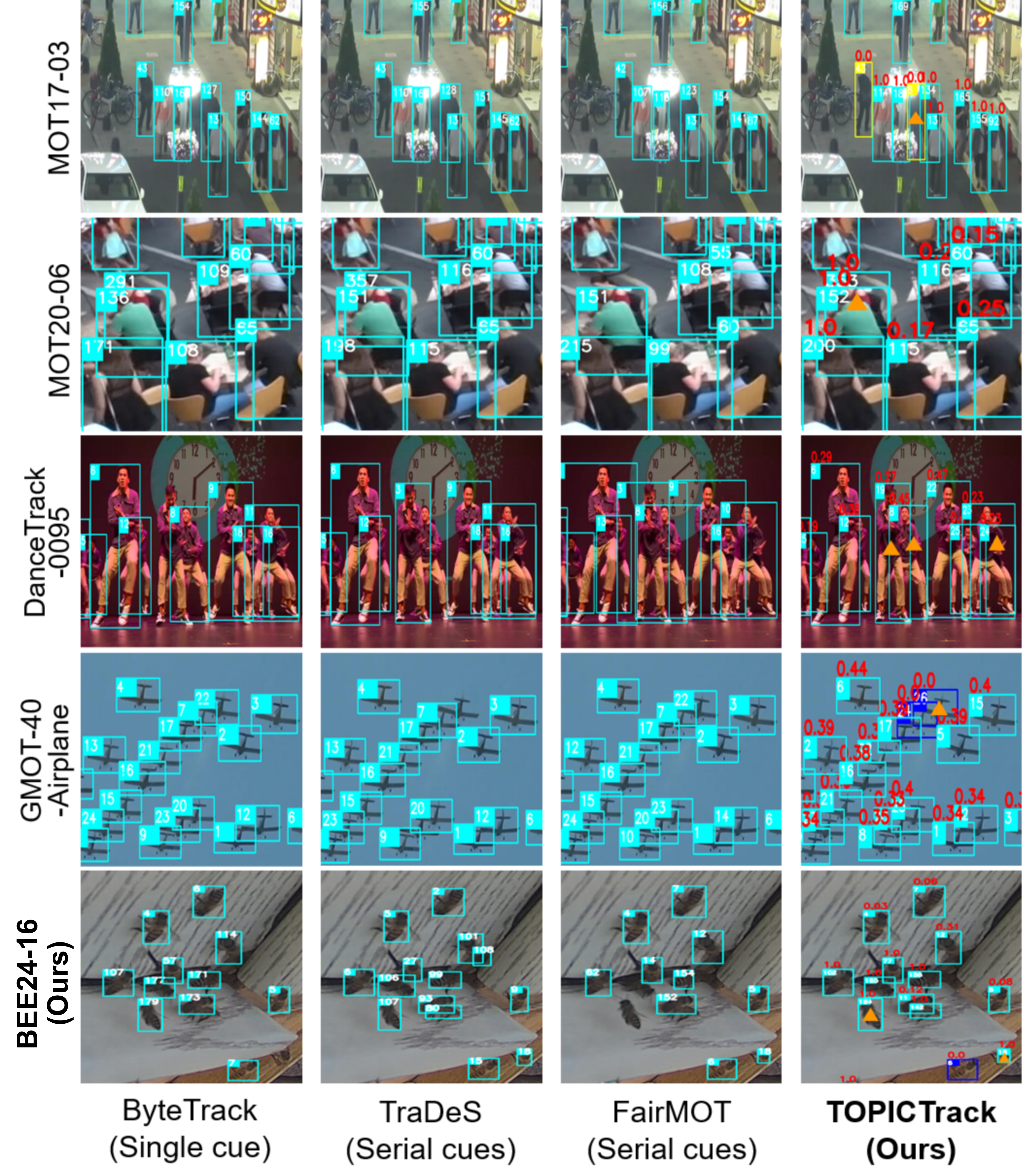}
    \vspace{-0.28cm}
    \caption{Visualizes tracking results of three state-of-the-art trackers with different association paradigms and our proposed TOPICTrack on five datasets.
    The \textcolor{cyan}{cyan} rectangular boxes indicate positions where objects are normally tracked.
    The \textcolor[RGB]{245, 245, 5}{yellow} and \textcolor{blue}{blue} rectangular boxes denote objects tracked by our method using appearance and motion features when encountering matching conflicts. 
    The white values in the top left corner of the boxes indicate objects' ID, while the \textcolor[RGB]{207, 3, 5}{red} values above the boxes in the last column indicate objects' motion levels.
    And the \textcolor{orange}{orange} triangles are used to emphasize instances where our method correctly matches but other trackers missed.}
    \label{fig:benchmark_visualization}
\end{figure}

Figure~\ref{fig:benchmark_visualization} visualizes several tracking results of three state-of-the-art trackers and our proposed TOPICTrack in test sets of five datasets (MOT17, MOT20, DanceTrack, GMOT-40, and BEE24).
In the figure, the cyan rectangular boxes indicate the positions where the objects are normally tracked.
The yellow and blue rectangular boxes represent the objects tracked by our method using appearance and motion features when encountering matching conflicts. 
The white values in the top left corner of the boxes indicate the object' ID, while the red values above the boxes in the last column of the figure indicate the object's motion level. 
The orange triangles are used to emphasize instances where our method correctly matches that other trackers missed.
According to Figure~\ref{fig:benchmark_visualization}, we have the following conclusions: 
(1) analyzing the results from MOT17-03, it becomes evident that our method successfully tracked a pedestrian with low motion intensity but a similar appearance. This is a task that neither motion nor appearance features alone managed to achieve in other algorithms. 
This illustrates the stronger capability of our approach in representing object appearances;
(2) from the results of MOT20-06, we observe that the TOPICTrack outperforms ByteTrack~\cite{zhang2022bytetrack} in accurately tracking pedestrians with rapid motion and high occlusion. 
This suggests that utilizing appearance features can make up for the limitations of motion features;
(3) from the results of DanceTrack-0095 and GMOT-40-Airplane, we can see that our approach better addresses challenges posed by highly similar appearances and severe occlusion, attributed to the utilization of motion features;
(4) from the results of BEE24-16, we note that the TOPICTrack outperforms purely motion-based trackers (ByteTrack~\cite{zhang2022bytetrack}) in successfully tracking rapidly flying bees, indicating that our proposed approach can better cope with the problem of complex motion patterns.

\begin{figure}[!h]
\setlength{\abovecaptionskip}{1pt}
\setlength{\belowcaptionskip}{1pt}
    \centering
    \includegraphics[width=0.95\linewidth]{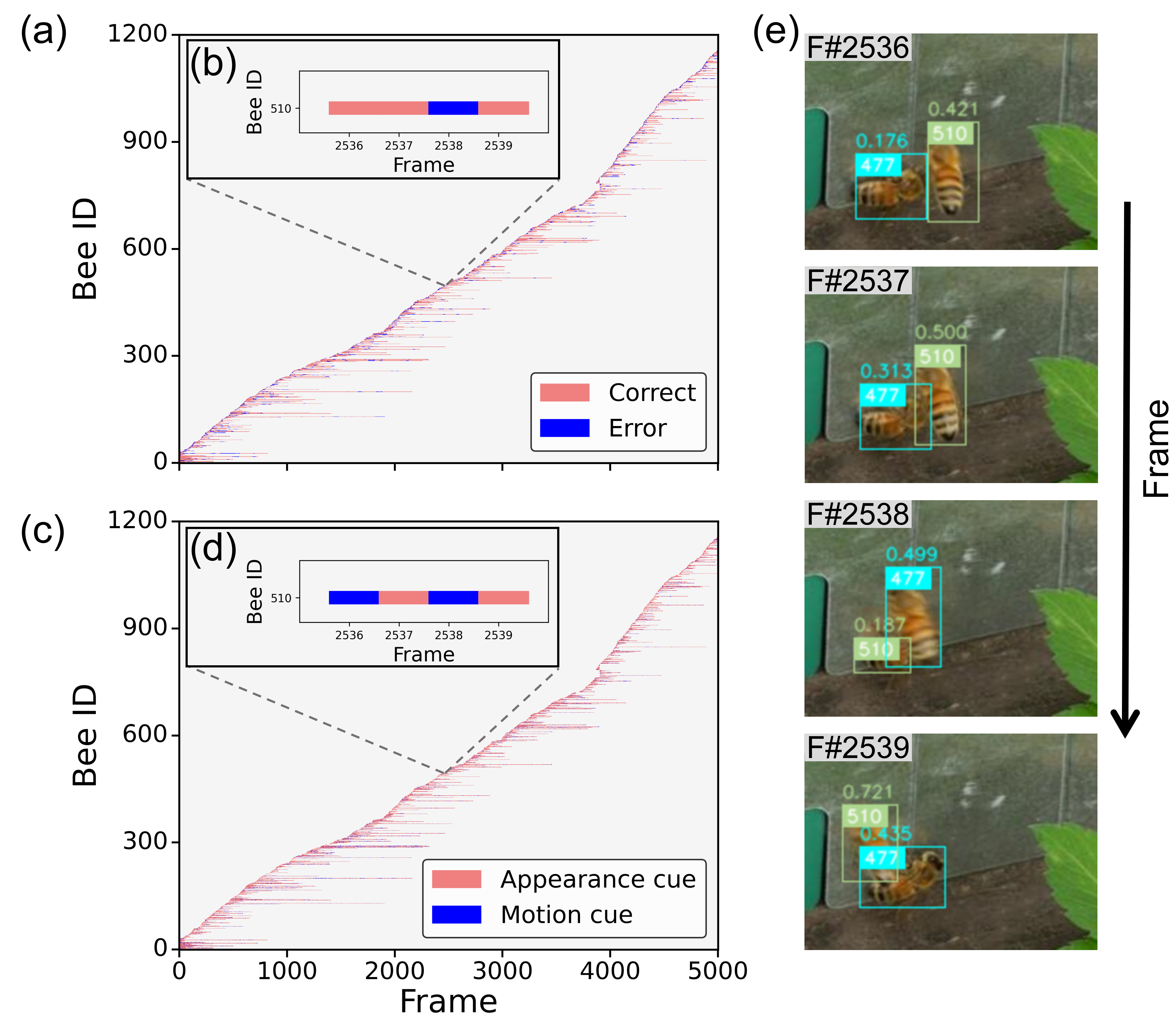}
    \vspace{-0.28cm}
    \caption{
    Accuracy and algorithmic strategy analysis for long-term tracking, using the BEE24-18 sequence as an example.
    (a) the tracking results (correct/error) for each bee, with red segments indicating correct tracking and blue segments representing tracking errors over the frames;
    (b) a zoomed-in view of (a) highlighting bee No.~510’s tracking results from frames 2536 to 2539; 
    (c) the primary assignment metric (appearance/motion) selected by TOPICTrack at each frame, where red segments indicate the usage of appearance cues and blue segments indicate motion cues.
    (d) a zoomed-in view of (c) focusing on bee No.~510’s cue selection  from frames 2536 to 2539;
    (e) tracking performance at corresponding times in the video with other subplots.
    }
    \label{fig:correct_error_with_strategy}
\end{figure}

To further demonstrate the effectiveness of our proposed method in long-period tracking scenarios, we conduct an in-depth analysis of the test dataset of BEE24. Statistically, our tracker's maximum number of consecutively tracked frames is 3,877 (about 3 minutes). In the following, we use a representative video sequence, BEE24-18, of over 5,000 frames. As shown in Figure~\ref{fig:correct_error_with_strategy}(a), it is evident that TOPICTrack maintains a high level of accuracy over most of the tracking period.
Figure~\ref{fig:correct_error_with_strategy}(b) and (d) present local magnifications of the tracking results and cue selection, illustrating the tracking process of bee No.~510 from frames 2536 to 2539. Figure~\ref{fig:correct_error_with_strategy}(e) displays the tracking performance in the video's corresponding period.
In frame 2536, the motion level of the bee No.~510 falls below the threshold $\alpha=0.5$. Consequently, TOPICTrack utilizes motion cues as the primary assignment metric, successfully maintaining the tracking of this bee. In frame 2537, the motion level of bee ID 510 reaches the $\alpha$, prompting the algorithm to automatically switch to appearance cues as the primary assignment metric, thereby continuing to track the bee successfully.
In frame 2538, the motion level of bee No.~510 drops below the $\alpha$, causing the algorithm to revert to using motion cues as the primary assignment metric. At this point, however, bees No.~477 and 510 are very close to each other, resulting in the motion model erroneously assigning trajectories to both bees and causing ID switches. This indicates the limitations of the adopted motion model.
In frame 2539, the motion level of the bee No.~510 suddenly increases, leading the association metric to primarily rely on appearance cues. At this point, the two bees are highly distinguishable in appearance. This helps TOPICTrack to accurately resume tracking bee No.~510.
This demonstrates that our proposed algorithm heuristically exploits motion levels to adaptively select the most appropriate association metrics at each instant, thereby providing the algorithm with automatic error correction capabilities and achieving long-term robust tracking.

\section{Conclusion}

In this paper, we propose a new MOT dataset called BEE24, which challenges models to track multiple similar-appearing small objects with complex motions over long periods. We believe that BEE24 contributes to the application and development of MOT techniques in the real world.
In terms of algorithm optimization, in order to take full advantage of the appearance and motion features, we propose a novel parallel association paradigm and present the TOPIC to implement it. 
The TOPIC can adaptively select the appearance or motion feature for association according to the motion level of objects.
Furthermore, we propose the AARM to enhance the tracker's ability to represent object appearance.
Exhaustive experiments demonstrate the effectiveness and superiority of our proposed tracker on five datasets.
From the algorithm side, we will consider optimizing the detection model and motion model to further improve the tracking performance.

\section*{Acknowledgments}
This work is supported by the National Natural Science Foundation of China (62072383), and the Fundamental Research Funds for the Central Universities (20720210044).

{
\bibliographystyle{IEEEtran}
\bibliography{ref}
}

\begin{IEEEbiography}[{\includegraphics[width=1in,height=1.25in,clip,keepaspectratio]{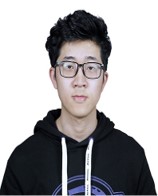}}]{Xiaoyan Cao}
received his Bachelor's degree from Jinan University in 2019 and his Master's degree from Xiamen University in 2022. He is currently a PhD candidate at the School of Environment and Energy, Peking University Shenzhen Graduate School. 

His main research interests include AI for science, urban flood prediction, and computer vision.
\end{IEEEbiography}

\vspace{-1cm}
\begin{IEEEbiography}
[{\includegraphics[width=1in,height=1.25in,clip,keepaspectratio]{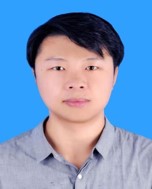}}]
{Yiyao Zheng}
received a Master's degree in Computer Technology Engineering from Huaqiao University in 2021. Currently, he works at Quanzhou Institute of Information Engineering and is pursuing a PhD in Computer Science at UITM Malaysia. 

His research areas include computer vision and digital twins.
\end{IEEEbiography}

\vspace{-1cm}
\begin{IEEEbiography}[{\includegraphics[width=1in,height=1.25in,clip,keepaspectratio]{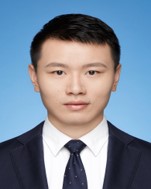}}]{Yao Yao}
received a bachelor’s degree from Beihang University, Beijing, China, in 2015, and a master’s degree from Tsinghua University, Beijing, China, in 2019. In January 2024, he earned a Ph.D. degree from the Tsinghua-Berkeley Shenzhen Institute (TBSI), Tsinghua University, Shenzhen, China. 
He is currently a senior research scientist at Jen Music AI, where his research focuses on AI-generated content and AI for Science. 

\end{IEEEbiography}

\vspace{-1cm}
\begin{IEEEbiography}[{\includegraphics[width=1in,height=1.25in,clip,keepaspectratio]{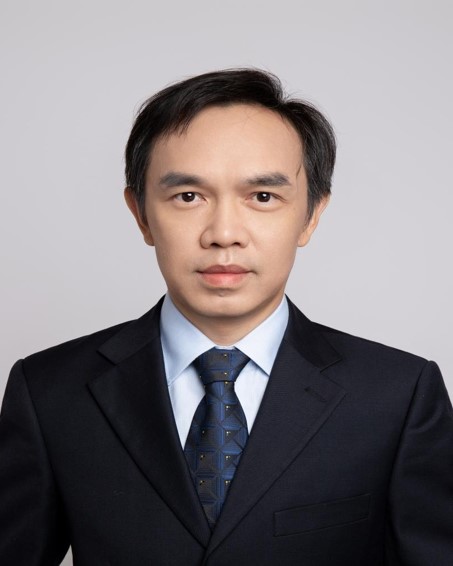}}]{Huapeng Qin}
received the B.S. degree from Tsinghua University, Beijing, China, in 1996, and the Ph.D. degree from Peking University, Beijing, China, in 2001.
He is currently a Professor at the School of
Environment and Energy, Peking University Shenzhen Graduate School, Shenzhen, China. 

His research interests include low-impact development and sponge cities, Intelligent water systems, and AI approaches that integrate data-driven and knowledge-driven methods.
\end{IEEEbiography}

\vspace{-1cm}
\begin{IEEEbiography}[{\includegraphics[width=1in,height=1.25in,clip,keepaspectratio]{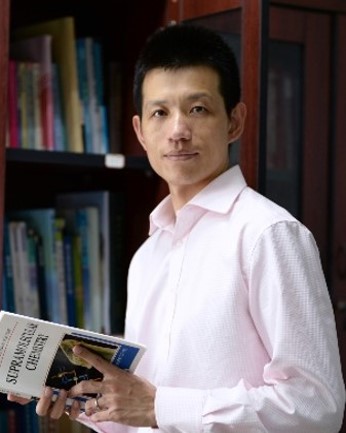}}]{Xiaoyu Cao}
received the Ph.D. from the University of Strasbourg, Strasbourg, France, in 2009, under Prof. Jean-Marie Lehn, a Nobel Laureate in Chemistry (1987).  
He is currently a Professor at Xiamen University. 

His research interests include supramolecular chemistry, dynamic covalent chemistry, host–guest chemistry, design and synthesis of novel conjugated organic materials for optoelectronic and biological applications.
\end{IEEEbiography}

\vspace{-1cm}
\begin{IEEEbiography}[{\includegraphics[width=1in,height=1.25in,clip,keepaspectratio]{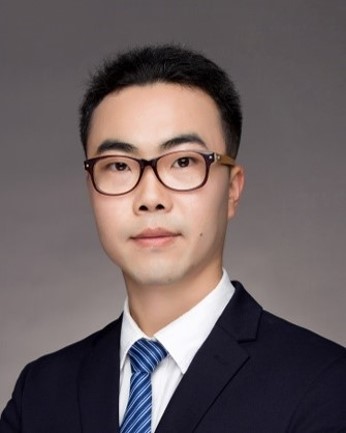}}]{Shihui Guo}
is a professor at School of Informatics, Xiamen University. He received his Ph.D. in computer animation from National Centre for Computer Animation, Bournemouth University, UK and B.s. in Electrical Engineering from Peking University, China. 

His research interests include computer graphics and vision, humancomputer interaction.
\end{IEEEbiography}

\vfill

\end{document}